# Efficient Multi-Start Strategies for Local Search Algorithms

**András György**                                                    GYA@SZIT.BME.HU
*Machine Learning Research Group*
*Computer and Automation Research Institute*
*of the Hungarian Academy of Sciences*
*1111 Budapest, Hungary*

**Levente Kocsis**                                                  KOCSIS@SZTAKI.HU
*Data Mining and Web Search Research Group, Informatics Laboratory*
*Computer and Automation Research Institute*
*of the Hungarian Academy of Sciences*
*1111 Budapest, Hungary*

## Abstract

Local search algorithms applied to optimization problems often suffer from getting trapped in a local optimum. The common solution for this deficiency is to restart the algorithm when no progress is observed. Alternatively, one can start multiple instances of a local search algorithm, and allocate computational resources (in particular, processing time) to the instances depending on their behavior. Hence, a multi-start strategy has to decide (dynamically) when to allocate additional resources to a particular instance and when to start new instances. In this paper we propose multi-start strategies motivated by works on multi-armed bandit problems and Lipschitz optimization with an unknown constant. The strategies continuously estimate the potential performance of each algorithm instance by supposing a convergence rate of the local search algorithm up to an unknown constant, and in every phase allocate resources to those instances that could converge to the optimum for a particular range of the constant. Asymptotic bounds are given on the performance of the strategies. In particular, we prove that at most a quadratic increase in the number of times the target function is evaluated is needed to achieve the performance of a local search algorithm started from the attraction region of the optimum. Experiments are provided using SPSA (Simultaneous Perturbation Stochastic Approximation) and k-means as local search algorithms, and the results indicate that the proposed strategies work well in practice, and, in all cases studied, need only logarithmically more evaluations of the target function as opposed to the theoretically suggested quadratic increase.

## 1. Introduction

Local search algorithms applied to optimization problems often suffer from getting trapped in a local optimum. Moreover, local search algorithms that are guaranteed to converge to a global optimum under some conditions (such as Simulated Annealing or Simultaneous Perturbation Stochastic Approximation, SPSA, see, e.g., Spall, Hill, & Stark, 2006), usually converge at a very slow pace when the conditions are satisfied. On the other hand, if the algorithms are employed with more aggressive settings, much faster convergence to local optima is achievable, but with no guarantee to find the global optimum. The common solu-tion to escape from a local optimum is to restart the algorithm when no progress is observed (see e.g., Martí, Moreno-Vega, & Duarte, 2010; Zabinsky, Bulger, & Khompatraporn, 2010, and the references therein).

Alternatively, one can start multiple instances of the local search algorithm, and allocate computational resources, in particular, processing time, to the instances depending on their behavior. Instances can be started at any time, and so the number of instances may grow over time depending on the allocation strategy. (see, e.g., Chapter 10 of Battiti, Brunato, & Mascia, 2008 and the references therein). In this type of problems the computational cost is usually measured as the total number of steps made by all search algorithm instances: this often reflects the situation that the evaluation of the target function to be optimized is expensive, and the costs related to determine which algorithms to use next are negligible compared to the former (e.g., this is clearly the case if the task is to tune the parameters of a system whose performance can only be tested via lengthy experiments, see, e.g., Bartz-Beielstein, 2006; Hutter, Hoos, Leyton-Brown, & Stützle, 2009). In this paper we address the above problem of dynamically starting several instances of local search algorithms and allocating resources to the instances based on their (potential) performance.

To our knowledge, solutions to the above problem have either been based on heuristics or on the assumption that the local optima the search algorithms converge to have an extreme value distribution (see Section 2 below). In this paper, we propose new multi-start strategies under very mild conditions on the target function, with attractive theoretical and practical properties: Supposing a convergence rate of the local search algorithms up to an unknown constant, our strategies continuously estimate the potential performance of each algorithm instance and in every phase allocate resources to those instances that could converge to the optimum for a particular range of the constant. The selection mechanism is analogous to the DIRECT algorithm (Jones, Perttunen, & Stuckman, 1993; Finkel & Kelley, 2004; Horn, 2006) for optimizing Lipschitz-functions with an unknown constant, where preference is given to rectangles that may contain the global optimum. The optimum within each rectangle is estimated in an optimistic way, and the estimate depends on the size of the rectangle. In our strategies we use the function describing the convergence rate of the local search algorithms in a similar way as the size of the rectangles are used in the DIRECT algorithm.

Since in the proposed multi-start strategies the potential performance of each local search algorithm is continuously estimated from the currently best value of the target function returned by that algorithm, our method is restricted to work with local search algorithms that return the best known value of the target function after each step. This is the case, for example, in certain meta-learning problems, where the goal is to find a good parameter setting of a learning algorithm. Here the search space is the parameter space of the learning algorithm, and one step of the local search methods means running the learning algorithm completely on a possibly very large data set. On the other hand, if the local search algorithm is some sort of a gradient search optimizing an error function over some training data, then the value of the target function is usually available only in the case of batch learning (potentially after some very cheap computations), but not when the gradient is estimated only from a few samples.

The rest of the paper is organized as follows. Section 2 summarizes related research. The problem is defined formally in Section 3. The new multi-start local search strategies of this paper are described and analyzed in Section 4: in Section 4.1 we deal with a selection mechanism among a fixed number of instances of the local search algorithm, while, in addition, simple schedules for starting new instances are also considered in Section 4.2, which are natural extensions of the case of finitely many local search algorithm instances. This section concludes with a discussion of the results in Section 4.3. Simulation results on real and synthetic data are provided in Section 5. Conclusions and future work are described in Section 6.

## 2. Related Work

The problem of allocating resources among several instances of search algorithms can be comfortably handled in a generalized version of the maximum $K$-armed bandit problem. The original version of this problem consists of several rounds, where in each round one chooses one of $K$ arms, receives some reward depending on the choice, with the goal of maximizing the highest reward received over several rounds. This model can easily be used for our problem by considering each local search algorithm instance as an arm: pulling an arm means taking one additional step of the corresponding algorithm, that is, evaluating the target function at a point suggested by that algorithm, and the reward received is the value of the target function at the sampled point. A generic algorithm for the standard maximum K-armed bandit problem, where each reward is assumed to have independent and identical distribution, is provided by Adam (2001), where the so-called reservation price of an instance is introduced, which gives the maximum amount of resources worth to spend on an instance: if an instance achieves its reservation price, it is useless to select it again. The computation of the reservation price depends on a model of the algorithm that can be learnt under some specific constraints.

Now consider a scenario where several instances of some, possibly randomized local search algorithms are run after each other with the goal of maximizing the expected performance. Each instance is run until it terminates. In this scenario it is natural to assume that the values returned by the instances (usually some local optima of the target function) are independent. Furthermore, since good search algorithms follow (usually heuristic) procedures that yield substantially better results than pure random guessing, Cicirello and Smith (2004, 2005) suggested that the rewards (evaluated target function values) of the search instances may be viewed as the maximum of many random variables (if the instances are run for sufficiently long time), and hence may be modeled by extreme value distributions. Several algorithms are based on this assumption, and are hence developed for the maximum $K$-armed bandit problem with returns following generalized extreme value distributions: Cicirello and Smith apply (somewhat heuristic) methods that use the above extreme-value-distribution assumption at each decision point of a meta-learning algorithm, while Streeter and Smith (2006a) use this model to obtain upper confidence bounds on the performance estimate of each type of algorithms used and then try only the algorithms with the best expected result. The latter is a theoretically justified example of the natural strategy to probe the algorithm instances for a while, estimate their future performance based on the results of this trial phase, and then use the most promising algorithm for the time remaining. Streeter and Smith (2006b) proposed a distribution free approach that combines a multi-armed bandit exploration strategy with a heuristic selection among the available arms.

While in the standard maximum $K$-armed bandit problem the rewards in each round are assumed to be independent, this is clearly not the case in our situation where the algorithm instances are run parallel and the reward for evaluating the target function at a point is the improvement upon the current maximum, since the samples chosen by a local search algorithm usually depend on previous samples. Nevertheless, the ideas and lessons learnt from the maximum $K$-armed bandit problems can be used in our case, as well: for example, the algorithm Threshold Ascent of Streeter and Smith (2006b) gives reasonably good solutions in our case, or the principle of probing instances for a while and then using the most promising in the time remaining also carries over to this situation easily: such algorithms, having first an exploration then an exploitation phase, will be referred to in the sequel as explore-and-exploit algorithms. In this class of algorithms, simple rules were suggested by Beck and Freuder (2004) to predict the future performance of each algorithm, while Carchrae and Beck (2004) employ Bayesian prediction.

Another related problem is to find fast algorithms among several ones that solve the same problem. More precisely, several algorithm instances are available that all produce the correct answer to a certain question if run for a sufficiently long time. The time needed for an algorithm instance to find the answer is assumed to be a random quantity with independent and identical distributions for all the instances, and the goal is to combine the given algorithms to minimize the expected running time until the answer is found. When the distribution of the running time is known, an optimal non-adaptive time-allocation strategy[1] is to perform a sequence of runs with a certain cut-off time that depends on the distribution (Luby, Sinclair, & Zuckerman, 1993). If the distribution is unknown, a particular running time sequence can be chosen that results in an expected total running time that is only a logarithmic factor larger than the optimum achievable if the distribution is known. We note that this strategy is among the few that provide a schedule that increases the number of algorithm instances. The above set-up can be specialized to our problem: the goal is to find an $\varepsilon$-optimal approximation of the optimum and the running time is the number of steps needed by the given search algorithm to achieve such an approximation. Note that in this case the running time of any algorithm instance providing an $\varepsilon$-suboptimal solution has to be defined to be infinity, but the results of Luby et al. remain valid if an $\varepsilon$-optimal solution can be found with positive probability. For the same problem, Kautz, Horvitz, Ruan, Gomes, and Selman (2002) proposed an allocation strategy based on updating dynamically the belief over the run-time distribution. Concerning the latter, Hoos and Stützle (1999) found empirically that run-time distributions are approximately exponential in certain (NP-hard) problems, while Ribeiro, Rosseti, and Vallejos (2009) dealt with the comparison of different run-time distributions.

Finally, when a set of time allocation strategies are available and the optimization problem is to be solved several times, one can use the standard multi-armed bandit framework as done by Gagliolo and Schmidhuber (2006, 2007, 2010).

Running several instances of an algorithm or several algorithms in parallel and selecting among the algorithms have been intensively studied, for example, in the area of meta-learning (Vilalta & Drissi, 2002) or automatic algorithm configuration (Hutter et al., 2009). The underlying problem is very similar in both cases: automatic algorithm configuration usually refers to tuning search algorithms, while meta-learning is used for a subset of these problems, tuning machine learning algorithms (the latter often allows more specific use of the data). The main problem here is to allocate time slices to particular algorithms with the aim of maximizing the best result returned. This allocation may depend on the intermediate performance of the algorithms. Most of the automatic algorithm configuration and meta-learning systems use various heuristics to explore the space of algorithms and parameters (see, e.g., Hutter et al., 2009).

---

1. In a non-adaptive time-allocation strategy the running time of an algorithm instance is fixed in advance, that is, the measured performance of the algorithm instances has no effect on the schedule.

Finally, it is important to note that, although multi-start local search strategies solve global optimization problems, we concentrate on maximizing the performance given the underlying family of local optimization methods. Since the choice of the latter has a major effect on the achievable performance, we do not compare our results to the vast literature on global optimization.

## 3. Preliminaries

Assume we wish to maximize a real valued function $f$ on the $d$-dimensional unit hypercube $[0, 1]^d$, that is, the goal is to find a maximizer $x^* \in [0, 1]^d$ such that $f(x^*) = f^*$ where

$$f^* = \max_{x \in [0,1]^d} f(x)$$

denotes the maximum of $f$ in $[0, 1]^d$. For simplicity, we assume that $f$ is continuous on $[0, 1]^d$.[2] The continuity of $f$ implies the existence of $x^*$, and, in particular, that $f$ is bounded. Therefore, without loss of generality, we assume that $f$ is non-negative.

If the form of $f$ is not known explicitly, search algorithms usually evaluate $f$ at several locations and return an estimate of $x^*$ and $f^*$ based on these observations. There is an obvious trade-off between the number of samples used (i.e., the number of points where the target function $f$ is evaluated) and the quality of the estimate, and the performance of any search strategy may be measured by the accuracy it achieves in estimating $f^*$ under a constraint on the number of samples used.

Given a local search algorithm $A$, a general strategy for finding a good approximation of the optimum $x^*$ is to run several instances of $A$ initialized at different starting points and approximate $f^*$ with the maximum $f$ value observed. We concentrate on local search algorithms $A$ defined formally by a sequence of possibly randomized sampling functions $s_n : [0, 1]^{d \cdot n} \to [0, 1]^d$, $n = 1, 2, \ldots$: $A$ evaluates $f$ at locations $X_1, X_2, \ldots$ where $X_{i+1} = s_i(X_1, \ldots, X_i)$ for $i \geq 1$, and the starting point $X_1 = s_0$ is chosen uniformly at random from $[0, 1]^d$; after $n$ observations $A$ returns the estimate of $x^*$ and the maximum $f^*$, respectively, by

$$\widehat{X}_n = \operatorname*{argmax}_{1 \leq k \leq n-1} f(X_k) \qquad \text{and} \qquad f(\widehat{X}_n).$$

where ties in the argmax function may be broken arbitrarily, that is, if more samples $X_k$ achieve the maximum, $\widehat{X}_n$ can be chosen to be any of them. To avoid ambiguity and simplify notation, here and in the following, unless stated explicitly otherwise, we adopt the convention to use argmax to denote the maximizing sample with the smallest index, but the results remain valid under any other choice to break ties.

---

2. The results can easily be extended to (arbitrary valued) bounded piecewise continuous functions with finitely many continuous components.

For simplicity, we consider only starting a single local search algorithm $A$ at different random points, although the results of this work can be extended to allow varying the parameters of $A$ (including the situation of running different local search algorithms, where a parameter would choose the actually employed search algorithm); as well as to allow dependence among the initializations of $A$ (that is, the starting point and parameters of a local search instance may depend on information previously obtained about the target function).

It is clear that if the starting points are sampled uniformly from $[0, 1]^d$ and each algorithm is evaluated at its starting point then this strategy is consistent, that is, $f(\widehat{X}_n)$ converges to the maximum of $f$ with probability 1 as the number of instances tends to infinity (in the worst case we perform a random search that is known to converge to the maximum almost surely). On the other hand, if algorithm $A$ has some favorable properties then it is possible to design multi-start strategies that still keep the random search based consistency, but provide much faster convergence to the optimum in terms of the number of evaluations of $f$.

Since the sequence $f(\widehat{X}_n)$ is bounded and non-decreasing, it converges (no matter what random effects occur during the search). The next lemma, proved in Appendix A, shows that, with high probability, the convergence cannot be arbitrarily slow.

**Lemma 1** *For any $\hat{f}^* \in [0, 1]^d$, let $E$ denote the event $\lim_{n\to\infty} f(\widehat{X}_n) = \hat{f}^*$.[3] If $P(E) > 0$, then, for any $0 < \delta < 1$ there is an event $E_\delta \subseteq E$ with $1 - P(E_\delta) < \delta$ such that $f(\widehat{X}_n) \to \hat{f}^*$ uniformly almost everywhere on $E_\delta$. In other words, there exists a non-negative, non-increasing function $g_\delta(n)$ with $\lim_{n\to\infty} g_\delta(n) = 0$ such that*

$$\mathbb{P}\left(\lim_{t\to\infty} f(\widehat{X}_t) - f(\widehat{X}_n) \leq g_\delta(n) \text{ for all } n \,\Big|\, \lim_{t\to\infty} f(\widehat{X}_t) = \hat{f}^*\right) \geq 1 - \delta \ . \tag{1}$$

In certain cases, $g_\delta(n) = O(e^{-\alpha n})$, as shown by Nesterov (2004) for (gradient-based) optimization for convex functions, by Gerencsér and Vágó (2001) for noise-free SPSA for convex functions, or by Kieffer (1982) for $k$-means clustering (or Lloyd's algorithm) in one dimension for log-concave densities. While these results pertain to the simple situation where there is only one local optimum which is the global one, many of these results can be extended to more general situations, and we observed exponential rate of convergence in our own experiments with functions with many local maxima.

The convergence property of local search algorithms guaranteed by Lemma 1 will be exploited in the next section to derive efficient multi-start search strategies.

## 4. Multi-Start Search Strategies

Standard multi-start search strategies run an instance of $A$ until it seems to converge to a location where there is no hope to beat the currently best approximation of $f^*$. An alternative way of using multiple instances of local search algorithms is to run all algorithms in parallel, and in each round decide which algorithms can take an extra step. This approach may be based on estimating the potential performance of a local search algorithm $A$ based on Lemma 1. Note that if $g_\delta$ were known, an obvious way would be to run each instance until their possible performances become separated with high probability in the sense that the margin between the performance of the actually best and the second best algorithm is so large that the actually best algorithm is guaranteed to be the best, in the long run, with high probability. Then we could just pick the best instance and run it until the given computational budget is exhausted (this would be a simple adaptation of the explore-and-exploit idea of choosing the best algorithm based on a trial phase as in Beck & Freuder, 2004; Carchrae & Beck, 2004).

---

3. In practice we can usually assume that local search algorithms converge to local optima, and so $\hat{f}^*$ may be assumed to be a local optimum.

In practice, however, $g_\delta$ is usually not known, but for certain problem classes and local search algorithms it may be known to belong to some function class, for example, $g_\delta$ may be known up to a (multiplicative) constant factor (here, for example, the constant may depend on certain characteristics of $f$, such as its maximum local steepness). Even in the latter case, the best instance still cannot be selected with high probability no matter how large the margin is (as $g_\delta$ may be arbitrarily large). However, using ideas from the general methodology for Lipschitz optimization with an unknown constant (Jones et al., 1993), we can get around this problem and estimate, in a certain optimistic way, the potential performance of each algorithm instance, and in each round we can step the most promising ones.

The main idea of the resulting strategy can be summarized as follows. Assume we have $K$ instances of an algorithm $A$, denoted by $A_1, \ldots, A_K$. Let $X_{i,n}, i = 1, \ldots, K$ denote the location at which $f$ is evaluated by $A_i$ at the $n$th time it can take a step, where $X_{i,1}$ is the starting point of $A_i$. The estimate of the location of the maximum by algorithm $A_i$ after $n$ samples (steps) is

$$\widehat{X}_{i,n} = \operatorname*{argmax}_{1 \le t \le n} f(X_{i,t})$$

and the maximum value of the function is estimated by $\hat{f}_{i,n} = f(\widehat{X}_{i,n})$.

For any $i$, let $\hat{f}_i = \lim_{n \to \infty} f_{i,n}$ denote the limiting estimate of the maximum of $f$ provided by $A_i$. Let $g_\delta$ be defined as in Lemma 1 for the largest of these values,

$$\hat{f}^* = \max_{i=1,\ldots,K} \hat{f}_i.$$

Since $\hat{f}^*$ is the best achievable estimate of the maximum of $f$ given the actual algorithms $A_1, \ldots, A_K$, $g_\delta$ gives a high probability convergence rate for those algorithms that provide the best estimate of the maximum in the long run (note that the assumption deals with each limiting estimate – usually a local maximum – separately, that is, here no assumption is made on algorithms whose limiting estimates are less than $\hat{f}^*$). Then, if $A_i$ evaluates $f$ at $n_{i,r}$ points by the end of the $r$th round and $A_i$ converges to the best achievable estimate $\hat{f}^*$, by Lemma 1 we have, with probability at least $1 - \delta$,

$$\hat{f}_i - \hat{f}_{i,n_{i,r}} \le g_\delta(n_{i,r}),$$

and so

$$\hat{f}_{i,n_{i,r}} + g_\delta(n_{i,r}) \qquad (2)$$

is an optimistic estimate of $\hat{f}^*$. If $A_i$ is suboptimal in the sense that $\lim_{n\to\infty} \hat{f}_{i,n} < \hat{f}^*$ then the above estimate is still optimistic if the rate of convergence is not slower than $g_\delta$, and pessimistic if the rate of convergence is slower than $g_\delta$. The latter is desirable in the sense that we have a negatively biased estimate on the expected performance of an algorithm that we do not want to use (we should not waste samples on suboptimal choices).

As in practice $g_\delta$ is usually not known exactly, the estimate (2) often cannot be constructed. On the other hand, if $g_\delta$ is known up to a constant factor then we can construct a family of estimates for all scales: Let $g'_\delta$ denote a normalized version of $g_\delta$ such that $g'_\delta(0) = 1$ and $g_\delta(n)/g'_\delta(n)$ is a constant for all $n$, and construct the family of estimates

$$\hat{f}_{i,n_{i,r}} + c g'_\delta(n_{i,r}) \tag{3}$$

where $c$ ranges over all positive reals. Then it is reasonable to choose, in each round, those algorithms to take another step that provide the largest estimate for some values of $c$ (typically, if an algorithm gives the largest estimate for some $c = c'$ then there is an interval $\mathcal{I}$ containing $c'$ such that the algorithm provides the largest estimate for any $c \in \mathcal{I}$). In this way we can get around the fact that we do not know the real scaling factor of $g'_\delta$, as we certainly use the algorithms that provide the largest value of (3) for $c = g_\delta(1)/g'_\delta(1)$, and, as it will be discussed later, we do not waste too many samples for algorithms that maximize (3) for other values of $c$. Using the optimistic estimate (3) is very similar, in spirit, to the optimistic estimates in the standard "upper confidence bound"-type solution to the multi-armed bandit problem (Auer, Cesa-Bianchi, & Fischer, 2002) or in the well-known $A^*$ search algorithm (Hart, Nilsson, & Raphael, 1968).

However, the exact (local) convergence rate is not known, even up to a constant factor, for many local search algorithms, and even if it is, the corresponding bounds are usually meaningful only in the asymptotic regime, which is often not of practical interest. Therefore, to give more freedom in the design of the algorithm, we are going to use an estimate of the form

$$\hat{f}_{i,n_{i,r}} + c h(n_{i,r}) \tag{4}$$

where, similarly to the requirements on $g_\delta$, $h$ is a positive, monotone decreasing function with $\lim_{n\to\infty} h(n) = 0$. We will also assume, without loss of generality, that $h(0) = 1$. The actual form of $h$ will be based on the theoretical analysis of the resulting algorithms and some heuristic considerations. Essentially we will use $h$ functions that converge to zero exponentially fast, which is in agreement with the exponentially fast local convergence rates in the examples given after Lemma 1. The optimal choice of $h$, given, for example, $g_\delta$, is not known, and is left for future work.

## 4.1 Constant Number of Instances

The above idea can be translated to the algorithm METAMAX(K) shown in Figure 1. Here we consider the case when we have a fixed number of instances, and our goal is to perform (almost) as well as the best of them (in hindsight), while using the minimum number of evaluations of $f$. Note the slight abuse of notation that in the METAMAX(K) algorithm $\widehat{X}_r$ and $\hat{f}_r$ denote the estimates of the algorithm after $r$ rounds (and not $r$ steps/samples).

In the first part of step (a) of METAMAX we sweep over all positive $\hat{c}$ and select local search algorithms that maximize the estimate (4). It is easy to see, that if $A_i$ maximizes

---

MetaMax(K): A multi-start strategy with $K$ algorithm instances.

**Parameters:** $K > 0$ and a positive, monotone decreasing function $h$ with $\lim_{n \to \infty} h(n) = 0$.

**Initialization:** For each $i = 1, \dots, K$, take a step with each algorithm $A_i$ once, and let $n_{i,0} = 1$ and $\hat{f}_{i,0} = f(X_{i,1})$.

For each round $r = 1, 2, \dots$

(a) For $i = 1, \dots, K$ select algorithm $A_i$ if there exists a $\hat{c} > 0$ such that

$$\hat{f}_{i,n_{i,r-1}} + \hat{c}h(n_{i,r-1}) > \hat{f}_{j,n_{j,r-1}} + \hat{c}h(n_{j,r-1}) \tag{5}$$

for all $j = 1, \dots, K$ such that $(n_{i,r-1}, \hat{f}_{i,n_{i,r-1}}) \neq (n_{j,r-1}, \hat{f}_{j,n_{j,r-1}})$. If there are several values of $i$ selected that have the same step number $n_{i,r-1}$ then keep only one of these selected uniformly at random.

(b) Step each selected $A_i$, and update variables. That is, set $n_{i,r} = n_{i,r-1} + 1$ if $A_i$ is selected, and $n_{i,r} = n_{i,r-1}$ otherwise. For each selected $A_i$ evaluate $f(X_{i,n_{i,r}})$ and compute the new estimates $\widehat{X}_{i,n_{i,r}}$ and $\hat{f}_{i,n_{i,r}}$.

(c) Let $I_r = \operatorname{argmax}_{i=1,\dots,K} \hat{f}_{i,n_{i,r}}$ denote the index of the algorithm with the currently largest estimate of $f^*$, and estimate the location of the maximum with $\widehat{X}_r = \widehat{X}_{I_r, n_{I_r,r}}$ and its value with $\hat{f}_r = \hat{f}_{I_r, n_{I_r,r}}$.

Figure 1: The MetaMax(K) algorithm.

(4) for a particular $\hat{c} = u$ then there is a closed interval $\mathcal{I}$ containing $u$ such that $A_i$ also maximizes (4) for any $\hat{c} \in \mathcal{I}$. Therefore, in each round, the strategy MetaMax(K) selects the local search algorithms $A_i$ for which the corresponding point $(h(n_{i,r-1}), \hat{f}_{i,n_{i,r-1}})$ is a corner of the upper convex hull of the set

$$\mathcal{P}_r = \{(h(n_{j,r-1}), \hat{f}_{j,n_{j,r-1}}) : j = 1, \dots, K\} \cup \{(0, \max_{1 \leq j \leq K} \hat{f}_{j,n_{j,r-1}})\}. \tag{6}$$

The selection mechanism is illustrated in Figure 2.

To avoid confusion, note that the random selection in step (a) of MetaMax(K) implies that if all algorithms are in exactly the same state, that is, $(n_{i,r-1}, \hat{f}_{i,n_{i,r-1}}) = (n_{j,r-1}, \hat{f}_{j,n_{j,r-1}})$ for all $i, j$, then one algorithm is selected uniformly at random (this pathological situation may arise, e.g., at the beginning of the algorithm or if all the local search algorithms give the same estimate of $f^*$ for some range of step numbers). Apart from the case when one of the least used algorithms provides the currently best estimate, which happens surely in the first round but usually does not happen later (and includes the previous pathological case), it is guaranteed that in each round we use at least two algorithms, one with the largest estimate $\hat{f}_{i,n_{i,r-1}} = \hat{f}_{r-1}$ (for very small values of $\hat{c}$), and one with the smallest step number $n_{j,r-1}$ (for very large values of $\hat{c}$). Thus, usually at most half of the total number of function calls to $f$ can be used by any optimal local search algorithm. This observation gives a practical lower bound (which is valid apart from the pathological situation mentioned above) on the proportion of function calls to $f$ made by optimal local search algorithms; surprisingly, Theorem 6 below shows that this lower bound is achieved by the algorithm asymptotically.

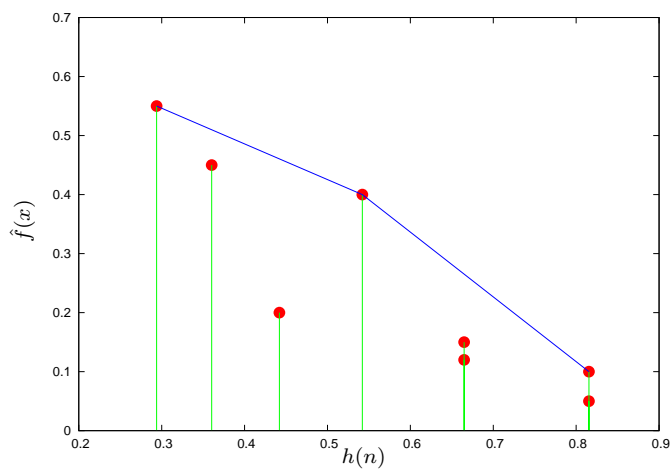

Figure 2: Selecting algorithm instances in METAMAX: the points represent the algorithm instances, and the algorithms that lie on the corners of the upper convex hull (drawn with blue lines) are selected.

The randomization in step (a) that precludes using multiple instances with the same step number is introduced to speed up the algorithm in certain pathological cases. For example, if $A_1$ converges to the correct estimate, while all the other algorithms $A_2, \ldots, A_K$ produce the same estimate in each round, independently of their samples, that is inferior to the estimates of $A_1$, then if we use the randomization, half of the calls to compute $f$ will be made by $A_1$, but without the randomization this would drop down to $1/K$ as in each round we would use each algorithm. Furthermore, we could take a step with all algorithms that lie on the convex hull, but similar pathological examples can be constructed when it is more beneficial to use only algorithms on the corners. On the other hand, it almost never happens in practice that three algorithms lie on the same line, and so algorithms typically never fall at non-corner points of the convex hull.

In the remainder of this section we analyze the performance of the METAMAX(K) algorithm. Proposition 2 shows that the algorithm is consistent in the sense that its performance asymptotically achieves that of the best algorithm instance as the number of rounds increases. To understand the algorithm better, Lemma 3 provides a general sufficient condition that an algorithm instance is not advanced in a given round, while, based on this result, Lemma 4 provides conditions that ensure that "suboptimal" algorithm instances are not used in a round if they have been stepped too many times (i.e., they have evaluated $f$ at too many points) before. Lemma 5 gives an upper bound on the number of algorithm instances used in a round. The results of the lemmas are then used to show in Theorems 6 and 8 and Remark 9 that "optimal" algorithm instances are used (asymptotically) at least at a minimum frequency that, in turn, yields the asymptotic rate of convergence of the METAMAX(K) algorithm.

The following proposition shows that the METAMAX(K) algorithm is consistent in a sense:

**Proposition 2** *The* METAMAX(K) *algorithm is consistent in the sense that* $\hat{f}_r \leq f^*$ *for all* $r$, *and*

$$f^* - \lim_{r \to \infty} \hat{f}_r = \min_{i=1,\ldots,K} \left\{ f^* - \lim_{n \to \infty} \hat{f}_{i,n} \right\}.$$

**Proof** The proof follows trivially from the fact that each algorithm is selected infinitely often, that is, $\lim_{r \to \infty} n_{i,r} = \infty$. To see the latter, we show that in every $K$ rounds the number of steps taken by the least used algorithm, that is, $\min_{i=1,\ldots,K} n_{i,r}$, is guaranteed to increase by one. That is, for all $k \geq 0$,

$$\min_{i=1,\ldots,K} n_{i,kK} \geq k. \tag{7}$$

As described above, in each round we select exactly one of the algorithms that have made the least number of steps. Thus, if there are $K$ such algorithms, the minimum step number per algorithm will increase in $K$ rounds, which completes the proof. $\square$

The METAMAX(K) algorithm is more efficient if suboptimal algorithms do not step too often. The next lemma provides sufficient conditions that an algorithm is not used in a given round.

**Lemma 3** *An algorithm instance* $A_j$ *is not used in any round* $r + 1$ *of the* METAMAX(K) *algorithm, if there are algorithms* $A_i$ *and* $A_k$ *such that* $\hat{f}_{i,n_{i,r}} > \hat{f}_{j,n_{j,r}} > \hat{f}_{k,n_{k,r}}$ *and either* $n_{i,r} \leq n_{j,r}$ *or*

$$\hat{f}_{j,n_{j,r}} \leq \hat{f}_{i,n_{i,r}} \left( 1 - \frac{h(n_{j,r})}{h(n_{k,r})} \right) + \hat{f}_{k,n_{k,r}} \frac{h(n_{j,r})}{h(n_{k,r})} \tag{8}$$

**Proof** As in each round the algorithms at the corners of the convex hull $\mathcal{P}_{r+1}$ are used, it is easy to see that an algorithm $A_j$ is not used in a round $r$ if there are algorithms $A_i$ and $A_k$ such that $\hat{f}_{i,n_{i,r}} > \hat{f}_{j,n_{j,r}} > \hat{f}_{k,n_{k,r}}$ and either $n_{i,r} \leq n_{j,r}$ or

$$\frac{\hat{f}_{i,n_{i,r}} - \hat{f}_{j,n_{j,r}}}{h(n_{j,r}) - h(n_{i,r})} \geq \frac{\hat{f}_{i,n_{i,r}} - \hat{f}_{k,n_{k,r}}}{h(n_{k,r}) - h(n_{i,r})}. \tag{9}$$

To finish the proof we show that (8) implies the latter. Indeed, (9) is equivalent to

$$h(n_{k,r})(\hat{f}_{i,n_{i,r}} - \hat{f}_{j,n_{j,r}}) \geq h(n_{j,r})(\hat{f}_{i,n_{i,r}} - \hat{f}_{k,n_{k,r}}) + h(n_{i,r})(\hat{f}_{k,n_{k,r}} - \hat{f}_{j,n_{j,r}}).$$

As the last term in the right hand side of the above inequality is negative by our assumptions, the inequality is satisfied if

$$h(n_{k,r})(\hat{f}_{i,n_{i,r}} - \hat{f}_{j,n_{j,r}}) \geq h(n_{j,r})(\hat{f}_{i,n_{i,r}} - \hat{f}_{k,n_{k,r}})$$

which is equivalent to (8). □

The above lemma provides conditions on not using some algorithm instances in a certain round that depend on the actual performance of the instances. The next result gives similar conditions, however, based on the best estimates (usually local optima) achievable with the algorithms. Let $\hat{f}_i^* = \lim_{r\to\infty} \hat{f}_{i,n_{i,r}}$ be the asymptotic estimate of algorithm $A_i$ for $f^*$, and let $\hat{f}^* = \max_{1 \le i \le K} \hat{f}_i^*$ denote the best estimate achievable using algorithms $A_1, \ldots, A_K$. Let $O \subseteq \{1, \ldots, K\}$ be the set of optimal algorithms that converge to the best estimate $\hat{f}^*$ (for these algorithms), and let $|O|$ denote the cardinality of $O$ (i.e., the number of optimal algorithm instances). Note that $O$ is a random variable that depends on the actual realizations of the possibly randomized search sequences of the algorithms. The next lemma shows that if $j \notin O$, then $A_j$ is not used at a round $r$ if it has been used too often so far.

**Lemma 4** *Let*

$$\Delta = \hat{f}^* - \max_{j \notin O} \hat{f}_j^*$$

*denote the margin between the estimates of the best and the second best algorithms. Then for any $0 < u < \Delta$ there is a random index $R^{(u)} > 0$ such that for any $j \notin O$, $A_j$ is not used by $\mathrm{MetaMax(K)}$ at a round $r + 1 > R^{(u)}$ if*

$$n_{j,r} \ge h^{-1} \left( h \left( \min_{i=1,\ldots,K} n_{i,r} \right) \left( 1 - \frac{\hat{f}_j^*}{\hat{f}^* - u} \right) \right). \tag{10}$$

*Furthermore, let $0 < \delta < 1$, for any $i \in O$, let $g_{\delta,i}$ denote the convergence rate of algorithm $A_i$ guaranteed by Lemma 1, and let $g_\delta(n) = \max_{i \in O} g_{\delta,i}(n)$ for all $n$. Then $P(R^{(u)} \le K g_\delta^{-1}(u) | \hat{f}_1^*, \ldots, \hat{f}_K^*) \ge 1 - \delta$ (where $g_\delta^{-1}$ is the generalized inverse of $g_\delta$), and no suboptimal algorithm $A_j, j \notin O$, is used for any $r > K g_\delta^{-1}(u)$ with probability at least $1 - \delta^{|O|}$ given the limiting estimates $\hat{f}_1^*, \ldots, \hat{f}_K^*$.*

**Proof** Let $i \in O$. Since $\lim_{n\to\infty} \hat{f}_{i,n} = \hat{f}_i^* = \hat{f}^*$ by assumption and (7) implies that $\lim_{r\to\infty} n_{i,r} = \infty$, there is an almost surely finite random index $R_i^{(u)} > 0$ such that for all $r > R_i^{(u)}$ we have $\hat{f}^* - \hat{f}_{i,n_{i,r}}^* \le u$ and so

$$\hat{f}^* - \hat{f}_r \le u. \tag{11}$$

Using Lemma 1 we can easily derive a high probability upper bound on $R^{(u)}$. Since for any $r > K g_\delta^{-1}(u) = R'$, (7) implies $n_{i,r} \ge g_\delta^{-1}(u)$, Lemma 1 yields

$$\mathbb{P} \left( \hat{f}_i^* - \hat{f}_{i,n_{i,r}} \le u \text{ for all } r > R' \mid \hat{f}_i^* = \hat{f}^* \right) \ge 1 - \delta.$$

It follows that with probability at least $1 - \delta^{|O|}$ there is an $i \in O$ such that $\hat{f}_i^* - \hat{f}_{i,n_{i,r}} \le u$, which implies that $R^{(u)}$ can be chosen such that $P(R^{(u)} > R' | \hat{f}_1^*, \ldots, \hat{f}_K^*) \le \delta^{|O|}$. Thus, to prove the lemma, it is enough to show (10).

Clearly, by (11), the algorithm will pick the estimate of one of the best algorithms after $R^{(u)}$ rounds. Let $A_k$ be an algorithm with the least number of steps taken up to the end of round $r$, that is, $n_{k,r} = \min_i n_{i,r}$. If $\hat{f}_{k,n_{k,r}} \geq \hat{f}_{j,n_{j,r}}$ then $A_j$ is not used in round $r+1$. Moreover, since $A_j \notin O$, $\hat{f}_{j,n_{j,r}} \leq \hat{f}_j^* < \hat{f}_r$ and in this case $A_j$ is not used in round $r+1$ if $n_{j,r} \geq n_{I_r,r}$ (recall that $n_{I_r,r}$ is the number of evaluations of $f$ initiated by the actually best algorithm $I_r$). Therefore, Lemma 3 implies that $A_j$ is not used for $r > R^{(\delta)}$ if

$$h(n_{j,r}) \leq h(n_{k,r}) \left( 1 - \frac{\hat{f}_{j,n_{j,r}} - \hat{f}_{k,n_{k,r}}}{\hat{f}_r - \hat{f}_{k,n_{k,r}}} \right).$$

This is clearly satisfied if

$$h(n_{j,r}) \leq h(\min_i n_{i,r}) \left( 1 - \frac{\hat{f}_j^*}{\hat{f}^* - u} \right) \tag{12}$$

for any $0 < u < \Delta$, since by (11) we have

$$1 - \frac{\hat{f}_{j,n_{j,r}} - \hat{f}_{k,n_{k,r}}}{\hat{f}_r - \hat{f}_{k,n_{k,r}}} \geq 1 - \frac{\hat{f}_{j,n_{j,r}}}{\hat{f}_r} \geq 1 - \frac{\hat{f}_j^*}{\hat{f}^* - u}.$$

Applying the inverse of $h$ to both sides of (12) proves (10), and, hence, the lemma. □

The above result provides an "individual" condition for each suboptimal algorithm for not being used in a round. On the other hand, if one of the optimal algorithms has been stepped sufficiently many times, we can give a cumulative upper bound on the number of suboptimal algorithms used in each round.

**Lemma 5** *Assume that $h$ decreases asymptotically at least exponentially fast, that is, there exist $0 < \alpha < 1$ and $n_\alpha > 0$ such that $\frac{h(n+1)}{h(n)} < \alpha$ for all $n > n_\alpha$. Assume that $r$ is large enough so that $n_{i,r} > n_\alpha$ for all $i$, and let $\varepsilon_r = 1 - \max_{i:\hat{f}_{i,n_{i,r}} \neq \hat{f}_r} \frac{\hat{f}_{i,n_{i,r}}}{\hat{f}_r} > 0$. Then at most $\lceil \frac{\ln \varepsilon_r}{\ln \alpha} \rceil + 1$ algorithms are stepped in round $r+1$, where $\lceil x \rceil$ denotes the smallest integer at least as large as $x$.*

**Proof** Let $i_0, i_1, \ldots, i_m$ denote the indices of the algorithms chosen in round $r+1$ with $\hat{f}_{i_0,r} < \hat{f}_{i_1,r} < \cdots < \hat{f}_{i_m,r} = \hat{f}_r$. Then Lemma 3 implies that $n_{i_0,r} < n_{i_1,r} < \cdots < n_{i_m,r}$ and

$$\hat{f}_r - \hat{f}_{i_k,r} < \alpha(\hat{f}_r - \hat{f}_{i_{k-1},r})$$

for all $k = 1, \ldots, m-1$. Repeated application of the above inequality implies

$$\hat{f}_r \varepsilon_r \leq \hat{f}_r - \hat{f}_{i_{m-1},r} < \alpha(\hat{f}_r - \hat{f}_{i_{m-2},r}) < \cdots < \alpha^{m-1}(\hat{f}_r - \hat{f}_{i_0,r}) \leq \hat{f}_r \alpha^{m-1}$$

which yields

$$m - 1 < \frac{\ln \varepsilon_r}{\ln \alpha}$$

As we assumed that $m + 1$ algorithms were chosen in round $r+1$, this fact finishes the proof. □

Based on Lemmas 4 and 5, the next theorem shows that if the local search algorithm converges fast enough (exponentially with a problem dependent rate, or faster than exponential) then half of the function calls to evaluate $f$ correspond to optimal algorithm instances.

**Theorem 6** *Assume that the performance of the algorithms $A_i, i = 1, \ldots, K$ are not all the same, that is, $|O| < K$, and suppose that*

$$\limsup_{n \to \infty} \frac{h(n+1)}{h(n)} < \min_{j \notin O} \left\{ 1 - \frac{\hat{f}_j^*}{\hat{f}^*} \right\}. \tag{13}$$

*Then asymptotically at least half of the function calls to evaluate $f$ in METAMAX(K) correspond to an optimal algorithm. That is,*

$$\mathbb{P} \left( \liminf_{r \to \infty} \frac{\sum_{i \in O} n_{i,r}}{\sum_{i=1}^K n_{i,r}} \geq \frac{1}{2} \ \Big| \ \hat{f}_1^*, \ldots, \hat{f}_K^* \right) = 1. \tag{14}$$

*Furthermore, for any $0 < \delta < 1$ and $\varepsilon > 0$ there is a constant $R^{(\delta,\varepsilon)} > 0$ such that*

$$\hat{f}^* - \hat{f}_r \leq g_\delta \left( \left\lfloor \frac{\sum_{i=1}^K n_{i,r}}{(2+\varepsilon)|O|} \right\rfloor \right) \tag{15}$$

*with probability at least $1 - |O|\delta$ given $\hat{f}_1^*, \ldots, \hat{f}_K^*$ simultaneously for all $r > R^{(\delta,\varepsilon)}$, where $g_\delta$ is defined in Lemma 4, and the threshold $R^{(\delta,\varepsilon)} > 0$ depends on $\delta, \varepsilon, h, g_\delta^{-1}$ and $\Delta$, where $g_\delta^{-1}$ and $\Delta$ are also defined in Lemma 4.*

**Proof** We show that a suboptimal $A_j$ is not chosen for large enough $r$ if $n_{j,r} > \min_k n_{k,r}$. By Lemma 4, it is sufficient to prove that, for large enough $r$,

$$\min_k n_{k,r} + 1 \geq h^{-1} \left( h(\min_k n_{k,r}) \left( 1 - \frac{\hat{f}_j^*}{\hat{f}^* - u} \right) \right) \tag{16}$$

for some $0 < u < \Delta$ (recall that here $r$ should be larger than $R^{(u)}$, the almost surely finite random index of Lemma 4).

As the minimum in (13) is taken over a finite set, it follows that there exists a small enough positive $u < \Delta$ such that

$$\limsup_{n \to \infty} \frac{h(n+1)}{h(n)} \leq \min_{j \notin O} \left\{ 1 - \frac{\hat{f}_j^*}{\hat{f}^* - u} \right\}, \tag{17}$$

which clearly implies (16) as $\lim_{r \to \infty} \min_k n_{k,r} = \infty$ by (7). This fact finishes the proof of (14), the first part of the theorem.

Next we prove (15). Let $N_u > 0$ be a threshold such that (17) holds for all $n \geq N_u$. Furthermore, by Lemma 1 and the union bound, (1) holds for each local search algorithm $A_i$ with $g_{\delta,i}$ in place of $g_\delta$ simultaneously for all $i \in O$ with probability at least $1 - |O|\delta$.

Then (7) and a slight modification of Lemma 4 imply that (16) holds simultaneously for all $r > K \max\{g_\delta^{-1}(u), N_u\} = R'$ with probability at least $1 - |O|\delta$ given $\hat{f}_1^*, \ldots, \hat{f}_K^*$. Since in each such round at most two algorithms are used, for any $r > R' + c$ we have $\frac{\sum_{i \in O} n_{i,r}}{\sum_{i=1}^K n_{i,r}} > \frac{c + R'}{2c + KR'}$ with high probability. Since the latter is bounded from below by $1/(2 + \varepsilon)$ for $c \geq R'(K - 2 - \varepsilon)/\varepsilon$, we have $\sum_{i \in O} n_{i,r} \geq \frac{\sum_{i=1}^K n_{i,r}}{2 + \varepsilon}$ for any $r > R^{(\delta, \varepsilon)} = R' + R'(K - 2 - \varepsilon)/\varepsilon$ with high probability. Then there is an algorithm $A_i, i \in O$ such that it is used in at least $\left\lceil \frac{\sum_{i=1}^K n_{i,r}}{|O|(2 + \varepsilon)} \right\rceil$ rounds, implying the statement of the theorem via Lemma 1. □

**Remark 7** *The above proof of Theorem 6 is based on Lemma 4. A proof based on Lemma 5 is also possible, since setting $\alpha = \min_{j \notin O}(1 - \hat{f}_j^*/\hat{f}^*)$ in the lemma, (13) implies that, for large enough $r$, $\epsilon_r \approx \alpha$, and so each round is approximately of length 2 by the lemma.*

It may happen that although the decay rate of $h$ is exponential, it is not quite fast enough to satisfy (13), so the optimal scenarios in the above theorem do not hold. In this case it turns out that the number of algorithms converging to the same local maximum plays the key role in determining the usage frequency of the optimal algorithms.

**Theorem 8** *Assume that the estimates provided by the algorithms $A_1, \ldots, A_K$ converge to $N > 1$ distinct limit points, such that $k_0 = |O|$ algorithms converge to $\hat{f}^*$, and $k_1, k_2, \ldots, k_{N-1}$ algorithms converge to each suboptimal limit points, respectively. Suppose furthermore that $h$ decreases asymptotically at least exponentially fast, that is, for some $0 < \alpha < 1$, $\limsup_{n \to \infty} \frac{h(n+1)}{h(n)} < \alpha$. Then*

$$\mathbb{P}\left(\liminf_{r \to \infty} \frac{\sum_{i \in O} n_{i,r}}{\sum_{i=1}^K n_{i,r}} \geq \frac{k_{\max}}{K - k_0 + k_{\max}} \,\middle|\, \hat{f}_1^*, \ldots, \hat{f}_K^*\right) = 1.$$

*where $k_{\max} = \max_{1 \leq i \leq N-1} k_i$.*

*Furthermore, using the definitions of Lemma 4 for any $0 < \delta < 1$ and $\varepsilon > 0$ there is a constant threshold $R^{(\delta, \varepsilon)}$*

$$\hat{f}^* - \hat{f}_r \leq \hat{c}\hat{g}_\delta\left(\left\lfloor \frac{k_{\max} \sum_{i=1}^K n_{i,r}}{(K - k_0 + k_{\max} + \varepsilon)|O|} \right\rfloor\right)$$

*with probability at least $1 - |O|\delta$ given $\hat{f}_1^*, \ldots, \hat{f}_K^*$ simultaneously for all $r > R^{(\delta, \varepsilon)}$, where $R^{(\delta, \varepsilon)}$ depends on $\varepsilon, \hat{f}_1^*, \ldots, sf_K$ and the convergence rate of all the algorithms[4].*

**Proof** Suppose $\hat{f}_1^*, \ldots, \hat{f}_K^*$ are given, and fix the random trajectories of all the algorithms. If there is a single suboptimal algorithm then the statement is trivial as $k_{\max}/(K - k_0 + k_{\max}) = 1/2$ and in each round at least two algorithms are used, and at most one of them can be the suboptimal one. From now on assume there are at least two suboptimal algorithms. Assume that $A_j$ and $A_k$ converge to *suboptimal* local maxima (strictly less than $\hat{f}^*$). For any $r$ large enough, an optimal algorithm $A_i$ is better than any of the suboptimal ones, that is, if $A_i$ converges to $\hat{f}^*$ then $\hat{f}_{i,n_{i,r}} > \hat{f}_{j,n_{j,r}}, \hat{f}_{k,n_{k,r}}$. Assume, without loss of generality, that $\hat{f}_{j,n_{j,r}} \geq \hat{f}_{k,n_{k,r}}$. If $n_{j,r} \leq n_{k,r}$ then clearly only $A_j$ can be chosen in round $r+1$. Assume $n_{j,r} > n_{k,r}$. Since $\hat{f}_{j,n_{j,r}}$ and $\hat{f}_{k,n_{k,r}}$ are convergent sequences (as $r \to \infty$), for large enough $r$ we have, for some $\alpha_{j,k} < 1$,

$$1 - \frac{\hat{f}_{j,n_{j,r}} - \hat{f}_{k,n_{k,r}}}{\hat{f}_{i,n_{i,r}} - \hat{f}_{k,n_{k,r}}} > \alpha_{j,k} \geq \alpha^{t_{j,k}} \tag{18}$$

where $t_{j,k} = \lceil \ln \alpha_{j,k}/\ln \alpha \rceil$ is a positive integer. Note that if $A_j$ and $A_k$ converge to the same point, that is, $\lim_{r\to\infty}(\hat{f}_{j,n_{j,r}} - \hat{f}_{k,n_{k,r}}) = 0$, the second term on the left hand side of (18) converges to 0, and so $\alpha_{j,k}$ can be chosen to be $\alpha$, implying $t_{j,k} = 1$. Rearranging the above inequality one obtains

$$(1 - \alpha^{t_{j,k}})\hat{f}_{i,n_{i,r}} + \alpha^{t_{j,k}}\hat{f}_{k,n_{k,r}} > \hat{f}_{j,n_{j,r}} . \tag{19}$$

---

4. Instead of the convergence rate of *all* algorithms, $R^{(\delta, \varepsilon)}$ may be defined to be dependent on $g_\delta$ and $\Delta$.

If $n_j - n_k \geq t_{j,k}$, then the conditions on $h$ and the fact that both $n_{j,r}$ and $n_{k,r}$ tend to infinity as $r \to \infty$ (recall (7)) imply that, for large enough $r$, $h(n_{j,r})/h(n_{k,r}) < \alpha^{t_{j,k}}$. Since $\hat{f}_{i,n_{i,r}} \geq \hat{f}_{k,n_{k,r}}$ for large enough $r$, from (19) we obtain

$$\hat{f}_{j,n_{j,r}} < (1 - \alpha^{t_{j,k}})\hat{f}_{i,ni,r} + \alpha^{t_{j,k}}\hat{f}_{k,n_{k,r}} \leq \left(1 - \frac{h(n_{j,r})}{h(n_{k,r})}\right)\hat{f}_{i,ni,r} + \frac{h(n_{j,r})}{h(n_{k,r})}\hat{f}_{k,n_{k,r}}.$$

Thus, by Lemma 3, if $r$ is large enough, $A_j$ cannot be used in round $r+1$ if $n_{j,r} - n_{k,r} \geq t_{j,k}$. Since both $n_{j,r}$ and $n_{k,r}$ tend to infinity, it follows that, for large enough $r$,

$$|n_{j,r} - n_{k,r}| \leq t_{j,k} \tag{20}$$

for any two suboptimal algorithms $A_j$ and $A_k$. Note that this fact also implies that from any set of suboptimal algorithms converging to the same point, eventually at most one can be used in any round (since the corresponding thresholds are $t_{j,k} = 1$).

Clearly, (7) implies that both $n_{j,r}$ and $n_{k,r}$ grow linearly with $r$, and since their differences is bounded by (20), $\lim_{r\to\infty} n_{j,r}/n_{k,r} = 1$. Therefore, for any suboptimal algorithm $A_j$, we have $\lim_{n\to\infty} n_{j,r}/r \leq 1/k_{\max}$ (this is the maximal rate of using elements from the largest group of suboptimal algorithms converging to the same local optimum). Finally, as an optimal algorithm is used in each round $r$, for large enough $r$, we have

$$\liminf_{r\to\infty} \frac{\sum_{i\in O} n_{i,r}}{\sum_{i=1}^{K} n_{i,r}} = \liminf_{r\to\infty} \frac{\sum_{i\in O} n_{i,r}}{\sum_{i\in O} n_{i,r} + \sum_{i\notin O} n_{i,r}}$$
$$\geq \lim_{r\to\infty} \frac{r}{r + (K - k_0)\frac{r}{k_{\max}}} = \frac{k_{\max}}{K - k_0 + k_{\max}} ,$$

where we used the fact that $a/(a+b)$ is an increasing function of $a$ for $a, b > 0$. Since the above inequality holds for all realizations of the trajectories of $A_1, \ldots, A_K$, given $\hat{f}_1^*, \ldots, \hat{f}_K^*$, the first statement of the theorem follows.

The second statement follows similarly to (15) in Theorem 6. Since the exact value of $R^{(\delta,\varepsilon)}$ is not of particular interest, the derivation is omitted. $\qquad \square$

**Remark 9** *The main message of the above theorem is the somewhat surprising observation that suboptimal algorithms are slowed down if there is a large group of suboptimal algorithms converging to the same local optimum; the rate suboptimal algorithms are used at is bounded by the the size of the largest such group.*

## 4.2 Unbounded Number of Instances

It is clear that if the local search algorithms are not consistent (i.e., they do not achieve the global optimum $f^*$), then, despite its favorable properties, the MetaMax(K) strategy is inconsistent, too. However, if we increase the number of algorithms to infinity then we get the consistency from random search, while still keeping the reasonably fast convergence rate from MetaMax(K).

Clearly, one needs to balance between exploration and exploitation, that is, we have to control how often we introduce a new algorithm. One solution is to let the MetaMax algorithm solve this problem: the MetaMax($\infty$) algorithm, given in Figure 3, is the extension of MetaMax(K) that is able to run with infinitely many local search algorithm instances. Here a major issue is that new local search algorithms have to be started from time to time (this ensures that the algorithm will converge to the global maximum of $f$ since it also performs a random search): this is implemented by modifying step (a) of the MetaMax(K) algorithm so that a new, randomly initialized local search algorithm is introduced in each round (randomly selecting one algorithm "uniformly" from the infinitely many possible algorithms not used so far). Obviously, we have to skip the initialization step of MetaMax(K) and "start" each algorithm with 0 samples. To better control the length of each round (i.e., the exploration), in each round $r$ we allow the use of a different function $h$, denoted by $h_{r-1}$ that may depend on any value measured before round $r$ (this is suppressed in the notation). As before, we assume that $h_r(0) = 1$, $h_r(n)$ is monotone decreasing in $n$, and $\lim_{n \to \infty} h_r(n) = 0$ for all $r$. Typically we will make $h_{r-1}$ be dependent on the total number of steps (i.e., the function calls to evaluate $f$) $t_{r-1} = \sum_{i=1}^{K_{r-1}} n_{i,r-1}$ made by all the algorithms before round $r$, where $K_{r-1}$ is the number of algorithm instances used before round $r$; note that $K_{r-1} = r - 1$ for all $r$, as we start exactly one new algorithm in each round.

It is desired that, although the number of local search algorithms grows to infinity, the number of times the best local search algorithm is advanced by the MetaMax($\infty$) algorithm approaches infinity reasonably fast. Somewhat relaxing the random initialization condition, we may imagine a situation where the local search algorithms are initialized in some clever, deterministic way, and so in the first few steps they do not find any better value than their initial guesses. If all algorithms are optimal (this may be viewed as a result of the clever initialization), then they may provide, for example, the identical estimates $0.5, 0.5, 1$ for the first three steps. Then it is easy to see that each algorithm is stepped exactly twice, thus no convergence to the optimum (which would be found after the third step) is achieved. Although the random initialization of the search algorithms guarantees the consistency of MetaMax($\infty$) (see Proposition 10 below), robust behavior in even such pathological cases is preferred.

This can be achieved by a slight modification of the algorithm: if in a round a local search algorithm overtakes the currently best algorithm, that is, $I_r \neq I_{r-1}$, then algorithm $A_{I_r}$

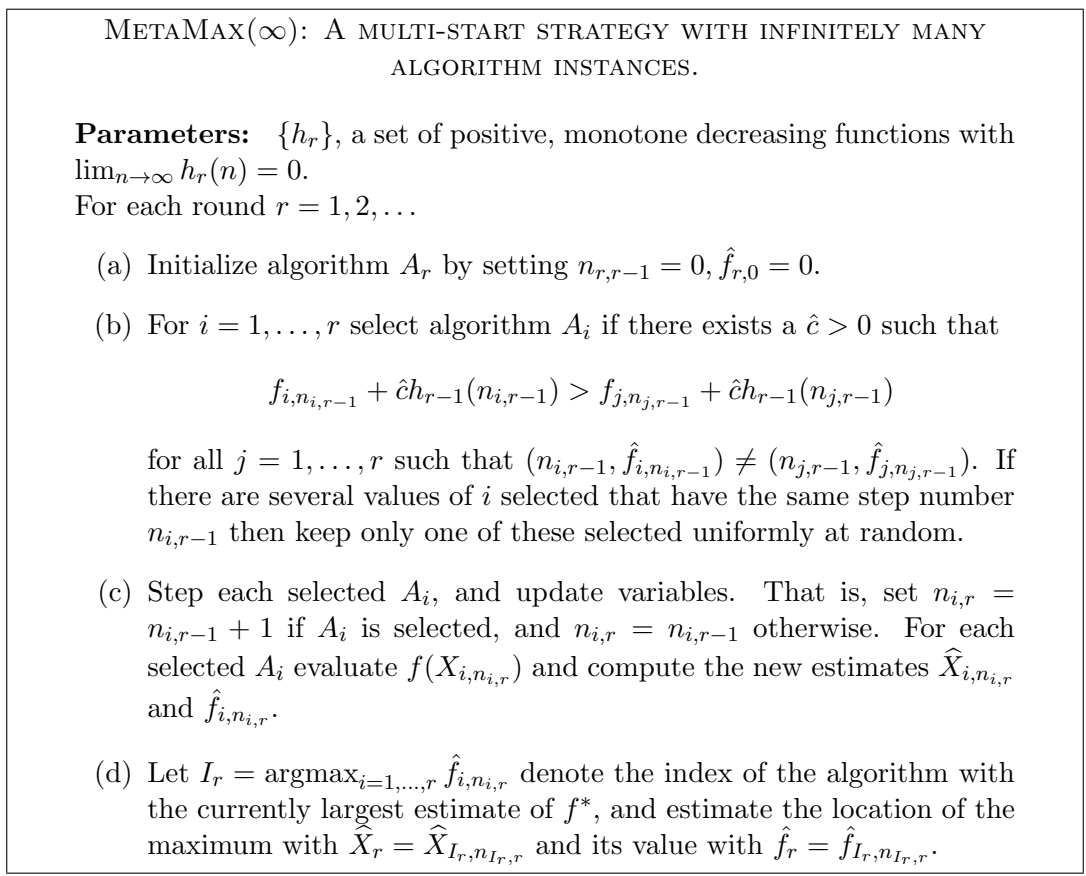

MetaMax($\infty$): A multi-start strategy with infinitely many algorithm instances.

**Parameters:** $\{h_r\}$, a set of positive, monotone decreasing functions with $\lim_{n\to\infty} h_r(n) = 0$.
For each round $r = 1, 2, \ldots$

(a) Initialize algorithm $A_r$ by setting $n_{r,r-1} = 0, \hat{f}_{r,0} = 0$.

(b) For $i = 1, \ldots, r$ select algorithm $A_i$ if there exists a $\hat{c} > 0$ such that

$$f_{i,n_{i,r-1}} + \hat{c} h_{r-1}(n_{i,r-1}) > f_{j,n_{j,r-1}} + \hat{c} h_{r-1}(n_{j,r-1})$$

for all $j = 1, \ldots, r$ such that $(n_{i,r-1}, \hat{f}_{i,n_{i,r-1}}) \neq (n_{j,r-1}, \hat{f}_{j,n_{j,r-1}})$. If there are several values of $i$ selected that have the same step number $n_{i,r-1}$ then keep only one of these selected uniformly at random.

(c) Step each selected $A_i$, and update variables. That is, set $n_{i,r} = n_{i,r-1} + 1$ if $A_i$ is selected, and $n_{i,r} = n_{i,r-1}$ otherwise. For each selected $A_i$ evaluate $f(X_{i,n_{i,r}})$ and compute the new estimates $\hat{X}_{i,n_{i,r}}$ and $\hat{f}_{i,n_{i,r}}$.

(d) Let $I_r = \text{argmax}_{i=1,\ldots,r} \hat{f}_{i,n_{i,r}}$ denote the index of the algorithm with the currently largest estimate of $f^*$, and estimate the location of the maximum with $\hat{X}_r = \hat{X}_{I_r,n_{I_r,r}}$ and its value with $\hat{f}_r = \hat{f}_{I_r,n_{I_r,r}}$.

Figure 3: The MetaMax($\infty$) algorithm.

is stepped several times until it is used more times than $A_{I_{r-1}}$.[5] The resulting algorithm, called MetaMax, is given in Figure 4. Note that the algorithms MetaMax($\infty$) and MetaMax conceptually differ only at one place: step (c) is extended with step (c') in the new algorithm. As a result, a technical modification also appears in step (d), and, to simplify the presentation of the MetaMax algorithm, a slight, insignificant modification is also introduced in step (b), see the discussion below.

The modification in MetaMax is not really significant in the practical examples we studied (see Section 5), as the number of steps taken by an algorithm that overtakes the currently best algorithm grows very quickly also in the MetaMax($\infty$) algorithm, since in MetaMax an overtake usually introduces very short rounds (close to the minimum length two in many cases) until the leading algorithm becomes also the most used one. The goal of the modification in step (b) is only to synchronize the choice of the optimal algorithms in steps (b) and (c'). An equally good solution would be to choose, in case of a tie in step

---

5. In this way we achieve that the actually best algorithm dominates all others in terms of both accuracy and the number of calls made by the algorithms to compute the target function. This is the same type of dominance as used by Hutter et al. (2009) in a slightly different context.

---

**METAMAX: A MULTI-START STRATEGY WITH INFINITELY MANY ALGORITHM INSTANCES.**

**Parameters:** $\{h_r\}$, a set of positive, monotone decreasing functions with $\lim_{n\to\infty} h_r(n) = 0$.

For each round $r = 1, 2, \ldots$

(a) Initialize algorithm $A_r$ by setting $n_{r,r-1} = 0, \hat{f}_{r,0} = 0$.

(b) For $i = 1, \ldots, r$ select algorithm $A_i$ if there exists a $\hat{c} > 0$ such that

$$f_{i,n_{i,r-1}} + \hat{c} h_{r-1}(n_{i,r-1}) > f_{j,n_{j,r-1}} + \hat{c} h_{r-1}(n_{j,r-1})$$

for all $j = 1, \ldots, r$ such that $(n_{i,r-1}, \hat{f}_{i,n_{i,r-1}}) \neq (n_{j,r-1}, \hat{f}_{j,n_{j,r-1}})$. If there are several values of $i$ selected that have the same step number $n_{i,r-1}$ then keep only one of these that has the smallest index.

(c) Step each selected $A_i$, and update variables. That is, set $n_{i,r} = n_{i,r-1} + 1$ if $A_i$ is selected, and $n_{i,r} = n_{i,r-1}$ otherwise. For each selected $A_i$ evaluate $f(X_{i,n_{i,r}})$ and compute the new estimates $\widehat{X}_{i,n_{i,r}}$ and $\hat{f}_{i,n_{i,r}}$.

(c') Let $I_r = \operatorname{argmax}_{i=1,\ldots,r} \hat{f}_{i,n_{i,r}}$ denote the index of the algorithm with the currently largest estimate of $f^*$ (in case $I_r$ is not unique, choose the one with the smallest number of steps $n_{i,r}$). If $I_r \neq I_{r-1}$, step algorithm $A_{I_r}$ $(n_{I_{r-1},r} - n_{I_r,r} + 1)$ times and set $n_{I_r,r} = n_{I_{r-1},r} + 1$.

(d) Estimate the location of the maximum with $\widehat{X}_r = \widehat{X}_{I_r,n_{I_r,r}}$ and its value with $\hat{f}_r = \hat{f}_{I_r,n_{I_r,r}}$.

Figure 4: The METAMAX algorithm.

(c'), an algorithm that was used in the current round. Also note that, as a result of the modifications, the currently best algorithm (with index $I_r$) has taken the most steps, and so the extra number of steps taken in step (c') is indeed positive. An important consequence of the modifications is that, in any round $r$, the number of steps taken by the local search algorithm $A_{I_r}$, which is the best at the end of the round, is between $r$ and $2r$ (see Theorem 15 below).

The rest of the section is devoted to the theoretical analysis of METAMAX($\infty$) and METAMAX, following the lines of the analysis provided for METAMAX(K). First, in Proposition 10, it is shown that the algorithm is consistent, that is, the solution found by the algorithm actually converges to $f^*$. Lemma 12 (a counterpart of Lemma 4) shows that suboptimal algorithms can make only finitely many steps, while Lemma 14 gives an upper bound on the length of each round. The main theoretical results of this section apply only to the METAMAX algorithm: Theorem 15 gives a lower bound on the number of steps taken by the actually best algorithm by the end of a given round, while, as a consequence, Theorem 16 shows the rate of convergence of the algorithm as a function of the total number of steps (i.e., function calls to evaluate $f$) used by all algorithm instances: it turns out that at most quadratically more steps are needed than for a generic local search algorithm instance that converges to the optimum.

Since the METAMAX($\infty$) and the METAMAX strategies perform a random search (the number of algorithms tends to infinity as the length of each round is finite), the algorithms are consistent:

**Proposition 10** *The strategies* METAMAX($\infty$) *and the* METAMAX *are consistent. That is,*

$$\lim_{r\to\infty} \hat{f}_r = f^*$$

*almost surely.*

**Proof** Clearly, the event that $\hat{f}_r$ does not converge to $f^*$ can be written as

$$\left\{\lim_{r\to\infty}\hat{f}_r \neq f^*\right\} = \bigcup_{n=1}^{\infty}\bigcap_{R=1}^{\infty}\bigcup_{r=R}^{\infty}\left\{\hat{f}_r < f^* - 1/n\right\} \tag{21}$$

Now the continuity of $f$ implies that, for any $n$, if $X$ is chosen uniformly from $[0,1]^d$ then $q_n = \mathbb{P}(f(X) > f^* - 1/n) > 0$. Thus, for any round $r$, $\mathbb{P}(\hat{f}_r < f^* - 1/n) \leq (1 - q_n)^r$, and so $\sum_{r=1}^{\infty}\mathbb{P}(\hat{f}_r < f^* - 1/n)$ is finite. Therefore, the Borel-Cantelli lemma (see, e.g., Ash & Doléans-Dade, 2000) implies that

$$\mathbb{P}\left(\bigcap_{R=1}^{\infty}\bigcup_{r=R}^{\infty}\left\{\hat{f}_r < f^* - 1/n\right\}\right) = 0$$

for all $n$. This, together with (21) finishes the proof, as

$$\mathbb{P}\left(\lim_{r\to\infty}\hat{f}_r \neq f^*\right) \leq \sum_{n=1}^{\infty}\mathbb{P}\left(\bigcap_{R=1}^{\infty}\bigcup_{r=R}^{\infty}\left\{\hat{f}_r < f^* - 1/n\right\}\right) = 0.$$

□

In the reminder of this section we will assume that local search algorithms that achieve an almost optimal value eventually converge to the optimum.

**Assumption 11** *Let* $\mathcal{F}^* \subset \mathbb{R}$ *denote the set of local maxima of* $f$, *and let* $\Delta = f^* - \sup_{\hat{f}\in\mathcal{F}^*, \hat{f}<f^*}\hat{f}$. *We assume that* $\Delta > 0$ *and if an algorithm* $A_i$ *is such that* $\hat{f}_{i,n} > f^* - \Delta$ *for some* $n$, *then* $\lim_{n\to\infty}\hat{f}_{i,n} = f^*$.

If the local search algorithms converge to local optima (which is a reasonable assumption in practice), the above assumption is usually satisfied: the only situation when it does not hold is the pathological case when $f$ has infinitely many local maxima and the set of these maxima is dense at the global maximum.

Under Assumption 11 we can prove, similarly to Lemma 4, that any suboptimal algorithm is selected only a limited number of times that increases with $h_r^{-1}$. In particular, if $h_r = h$ for all $r$ large enough, then any suboptimal algorithm is chosen only finitely many times.

**Lemma 12** *Suppose Assumption 11, and let $q = \mathbb{P}(f(X) > f^* - \Delta/2)$ for an $X$ uniformly distributed in $[0,1]^d$. Then, for both the MetaMax($\infty$) and the MetaMax algorithms, a suboptimal algorithm $A_j$ started before round $r+1$ is not used at round $r+1$, with probability at least $1 - (1-q)^r$, if*

$$n_{j,r} \geq h_r^{-1}\left(\frac{\Delta}{2f^* - \Delta}\right).$$

*In addition, if $h_r(n)$ is a non-decreasing function of $r$ for all $n$, then*

$$\limsup_{r \to \infty} \frac{n_{j,r}}{h_{r-1}^{-1}\left(\frac{\Delta}{2f^* - \Delta}\right)} < \infty \quad \text{almost surely.} \tag{22}$$

*In particular, if $h_r$ is a constant function, that is, $h_r = h_0$ for all $r$, then $\lim_{r \to \infty} n_{j,r} < \infty$ almost surely.*

**Remark 13** *Note that in the second part of the lemma we could drop the monotonicity assumption on $h_r$ and replace $h_{r-1}^{-1}\left(\frac{\Delta}{2f^* - \Delta}\right)$ with $\max_{0 \leq r' \leq r-1} h_{r'}^{-1}\left(\frac{\Delta}{2f^* - \Delta}\right)$ in (22).*

**Proof** Consider if algorithm $A_j$ is used at a round $r+1$. First note that with probability at least $1 - (1-q)^r$, $\hat{f}_r > f^* - \Delta/2$. Furthermore, the newly introduced algorithm, $A_{r+1}$ is not used yet, and we have $n_{r+1,r} = 0$ and $\hat{f}_{r+1,0} = 0$. Thus, by Lemma 3, $A_j$ is not used if

$$\hat{f}_{j,n_{j,r}} \leq \hat{f}_r\left(1 - \frac{h(n_{j,r})}{h_r(0)}\right) = \hat{f}_r\left(1 - h_r(n_{j,r})\right).$$

Since this is equivalent to

$$n_{j,r} \geq h_r^{-1}\left(1 - \frac{\hat{f}_{j,n_{j,r}}}{\hat{f}_r}\right),$$

and

$$h_r^{-1}\left(1 - \frac{\hat{f}_{j,n_{j,r}}}{\hat{f}_r}\right) \leq h_r^{-1}\left(\frac{\Delta}{2f^* - \Delta}\right)$$

by

$$\frac{\hat{f}_r - \hat{f}_{j,n_{j,r}}}{\hat{f}_r} \geq \frac{\hat{f}_r - (f^* - \Delta)}{\hat{f}_r} > \frac{(f^* - \Delta/2) - (f^* - \Delta)}{f^* - \Delta/2} = \frac{\Delta}{2f^* - \Delta}, \tag{23}$$

the first statement of the proof follows.

To prove the second part, let $\widehat{R}$ denote the first round in which there is an optimal algorithm $A_i$ with $\hat{f}_{i,n_{i,\widehat{R}}} > f^* - \Delta/2$. Then for any suboptimal algorithm $A_j$, the first part of the lemma implies that, for any $r > \widehat{R}$,

$$n_{j,r} \leq \max\left\{\widehat{R}, \max_{0 \leq r' \leq r-1} h_{r'}^{-1}\left(\frac{\Delta}{2f^* - \Delta}\right) + 1\right\} = \max\left\{\widehat{R}, \ h_{r-1}^{-1}\left(\frac{\Delta}{2f^* - \Delta}\right) + 1\right\}$$

where the equality holds since $h_r^{-1}(n)$ is non-decreasing in $r$. Thus

$$\limsup_{r\to\infty} \frac{n_{j,r}}{h_{r-1}^{-1}\left(\frac{\Delta}{2f^*-\Delta}\right)} \leq \limsup_{r\to\infty} \max\left\{ \frac{\widehat{R}}{h_{r-1}^{-1}\left(\frac{\Delta}{2f^*-\Delta}\right)},\ 1 + \frac{1}{h_{r-1}^{-1}\left(\frac{\Delta}{2f^*-\Delta}\right)} \right\}$$

$$\leq \max\left\{ \frac{\widehat{R}}{h_0^{-1}\left(\frac{\Delta}{2f^*-\Delta}\right)},\ 1 + \frac{1}{h_0^{-1}\left(\frac{\Delta}{2f^*-\Delta}\right)} \right\} \qquad (24)$$

where we used that $h_r^{-1}$ is non-decreasing in $r$. Since $\widehat{R}$ is finite, (24) is also finite with probability 1. □

A simple modification of Lemma 5 implies that once a $\Delta/2$-optimal sample point is found then only a limited number of suboptimal algorithms is chosen in each round.

**Lemma 14** *Consider algorithms* METAMAX($\infty$) *and* METAMAX. *Suppose Assumption 11 holds, and assume that $f^* - \hat{f}_R < \Delta/2$ for some $R > 0$. In any round $r > R$, if $h_r(n) = \alpha_r^n$ for some $0 < \alpha_r < 1$ and for all $n \geq 0$, then at most $\left\lceil \frac{\ln\frac{2f^*-\Delta}{\Delta}}{\ln(1/\alpha_r)} \right\rceil$ algorithms are chosen that have estimates $\hat{f}_j \leq f^* - \Delta$.*

**Proof** The proof follows from Lemma 5 taking into account that any suboptimal algorithm $A_j$ satisfies $\hat{f}_j^* \leq f^* - \Delta$ and that at least one optimal algorithm is chosen in each round $r > R$: Similarly to (23), $\varepsilon_r$ defined in Lemma 5 can be bounded as $\varepsilon_r > \Delta/(2f^*-\Delta)$, and so the number of suboptimal algorithms used in round $r$ is bounded by $\left\lceil \frac{\ln(1/\varepsilon_r)}{\ln(1/\alpha_r)} \right\rceil \leq \left\lceil \frac{\ln\frac{2f^*-\Delta}{\Delta}}{\ln(1/\alpha_r)} \right\rceil$. □

Finally we can derive the convergence rate of the algorithm METAMAX. First we bound the number of steps taken by the currently best algorithm, both in terms of the number of rounds and the total number of steps taken by all the local search algorithms.

**Theorem 15** *Consider the* METAMAX *algorithm. At the end of any round $r$ the number of steps taken by the currently best algorithm is between $r$ and $2r$. That is,*

$$r \leq n_{I_r,r} < 2r. \qquad (25)$$

*Furthermore, the number of calls $n_{I_r,r}$ to evaluate $f$ by the currently best algorithm $A_{I_r}$ can be bounded by a function of the total number of times $t_r = \sum_{i=1}^r n_{i,r}$ the target function $f$ is evaluated by all local search instances as*

$$n_{I_r,r} \geq \frac{\sqrt{2t_r + 7} - 1}{2}. \qquad (26)$$

**Proof** The first statement of the lemma is very simple, since in any round the actually best algorithm takes one step if there is no overtaking, and one or two steps if there is overtaking. Indeed, in any round $r \geq 2$, if there is no overtaking, that is, $I_r = I_{r-1}$, then $n_{I_r, r} = n_{I_r, r-1} + 1$. Otherwise, if $I_r \neq I_{r-1}$, then $n_{I_r, r} = n_{I_{r-1}, r} + 1$, and since $0 \leq n_{I_{r-1}, r} - n_{I_{r-1}, r-1} \leq 1$, we have

$$1 \leq n_{I_r, r} - n_{I_{r-1}, r-1} \leq 2$$

in all situations. Since in the first round clearly the only algorithm used takes 1 step, that is, $n_{I_1, 1} = 1$, (25) follows.

To prove the second part, notice that in any round $r$, at most $n_{I_{r-1}, r-1} + 1$ algorithms can be stepped in step $(c)$ as no algorithm can be used that has taken more steps than the currently best one. Also, in step $(c')$ no extra samples are used if there is no overtaking. In case of overtaking, $A_{I_r}$ has to be advanced in step $(c)$, as well as $A_{I_{r-1}}$, and so at most $n_{I_{r-1}, r-1} + 1$ extra steps has to be taken by $A_{I_r}$. Therefore,

$$t_r \leq t_{r-1} + 2 n_{I_{r-1}, r-1} + 2.$$

Thus, since no overtaking happens in round 1, we obtain

$$t_r \leq 1 + \sum_{s=2}^{r} 2(n_{I_{s-1}, s-1} + 1).$$

Then, by (25) we have

$$t_r \leq 1 + 4 \sum_{s=2}^{r} s = 1 + 2(r+2)(r-1) \leq 1 + 2(n_{I_r, r} + 2)(n_{I_r, r} - 1)$$

which yields (26). □

Note that in the proof we used a crude estimate on the length of a usual round (without overtaking) relative to, for example, Lemma 14. This, however, affects the result only by a constant factor as long as we are not able to bound the number of rounds or the number of extra steps taken when overtaking happens, since the effect of the overtakings itself introduces the quadratic dependence in the proof of (26). Experimental results in Section 5 show (see Figure 10) that the number of algorithm instances (which is in turn the number $r$ of rounds) has a usual growth rate of $\Omega(t_r / \ln t_r)$, which, if taken into account, may sharpen the bound on how often the best algorithm is chosen.

Under Assumption 11, the random search component of METAMAX implies that eventually we will have an optimal algorithm that is the best. From that point the convergence rate of the optimal local search algorithms determine the performance of the search, and the number of steps taken by the best local search algorithm is bounded by Theorem 15.

**Theorem 16** *Suppose Assumption 11 holds. Then there is an almost surely finite random index $R$ such that for all rounds $r > R$, the estimate $\hat{f}_r$ of the METAMAX algorithm and the total number of steps $t_r$ taken by all local search algorithms up to the end of round $r$ satisfies*

$$f^* - \hat{f}_r \leq g_\delta \left( \frac{\sqrt{2 t_r + 7} - 1}{2} \right)$$

*with probability at least $1 - \delta$, where $g_\delta$ is defined by Lemma 1 for the global maximum $f^*$.*

**Remark 17** *(i) The value of $R$ can be bounded with high probability using the properties of uniform random search for the actual problem; this would yield similar bounds as in Theorems 6 and 8 for the METAMAX(K) algorithm. (ii) Note the exploration-exploitation trade-off in the METAMAX algorithm: the value of $R$ is potentially decreased if we introduce new algorithms more often, while $n_{I_r,r}$ is reduced at the same time. (iii) Theorems 15 and 16 imply that, asymptotically, the METAMAX algorithm needs only quadratically more function evaluations than a local search algorithm that is ensured to converge to the optimum. In particular, if $f$ is of the form $f(x) = \sum_{i=1}^{s} f_i(x) \mathbb{I}_{S_i}(x)$ where the $S_i$ form a partition of $[0,1]^d$, $\mathbb{I}_{S_i}$ denotes the indicator function of $S_i$, and the $f_i$ belong to some nicely behaving function class such that a local search algorithm started in $S_i$ converges to the maximum of $f_i$ on $S_i$ (e.g., $f$ is a piecewise concave function with exponential convergence rate for the SPSA algorithm, which is used with a sufficiently small step size), then we can preserve the performance of a local search algorithm for the original function class at the price of an asymptotically quadratic increase in the number of function calls to evaluate $f$ (i.e., the total number of steps taken by all local search algorithm instances).*

## 4.3 Discussion on the Results

In some sense the theoretical results presented in the previous sections are weak. The consistency result for the METAMAX(K) algorithm follows easily from the fact that each local search algorithm is used infinitely many times, and the consistency of METAMAX($\infty$) and METAMAX follows from the consistency of a random search. The performance bounds provided have a disadvantage that they are asymptotic in the sense that they hold only after a possibly large number of rounds (a weakness of the bounds is that the minimum number of rounds is obtained from properties of uniform random search/sampling for the particular problem, neglecting more attractive properties of the algorithms). In fact, it is quite easy to construct scheduling strategies that are consistent and asymptotically an arbitrarily large fraction of the function evaluations (even almost all) is used by optimal local search algorithms: the explore-and-exploit algorithms achieve both of these goals if the number of function evaluations to be used is known ahead and they use an arbitrarily small fraction of the evaluations of the target function $f$ for exploration. We compare the performance of our algorithms to such explore-and-exploit algorithms in Section 5. In particular, to match the performance guarantees for the METAMAX family, we use algorithms that spend half of their time with exploration and half with exploitation, where in the exploration part a uniform allocation strategy is used when there is a finite number of local search algorithms, and the schedule of Luby et al. (1993) is used for infinitely many local search algorithms. Although the theoretical guarantees proved in this paper for the METAMAX family also hold for these explore-and-exploit algorithms, in all experiments the METAMAX family seems to behave superior compared to these algorithms, as expected.

The theoretical results also do not give sufficient guidance on how to chose the parameter $h$ or $h_r$ (the time-varying version of $h$ is not considered for the METAMAX(K) algorithm for simplicity and ease of presentation). Most of our results require a sufficiently fast exponential decay in $h$, which is problem dependent and cannot be determined in advance. A sufficiently fast decay rate would ensure, for example, that for the METAMAX(K) algorithm we could always use the stronger results of Theorem 6 and would never have to deal with the case when only the bound of Theorem 8 holds. One may easily choose $h$ to be any function that decreases super-exponentially: this would make the asymptotic bounds work, however, would slow down exploration (in the extreme case of $h_r(n) \equiv 0$, which is excluded by our conditions, no exploration would be performed, and the algorithms would use only the actually best local search algorithm). In practice we have always found that it is appropriate to chose $h_r$ decay exponentially. Furthermore, we have found it even more effective to gradually decrease the decay rate to enhance exploration as time elapses (the rationale behind this approach is the assumption that good algorithms should have more or less converged after a while, and so there may be a greater potential in exploration to improve our estimates). Finally, the connection between $g_\delta$ and $h$ should be further investigated.

Keeping the above limitations of the theoretical results in mind, we still believe that the theoretical analyses given provide important insight to the algorithms and may guide a potential user in practical applications, especially since the properties of the MetaMax family that can be proved in the asymptotic regime (e.g., that the rounds are quite short) can usually be observed in practice, as well. Furthermore, we think that it is possible to improve the analysis to bound the thresholds from which the results become valid with reasonable values, but this would require a different approach and, therefore, is left for future work.

## 5. Experiments

The variants of the MetaMax algorithm are tested in synthetic and real examples. Since there was only negligible difference in the performance of MetaMax($\infty$) and MetaMax,[6] in the following we present results only for MetaMax(K) and MetaMax. First we demonstrate the performance of our algorithm in optimizing a synthetic function (using SPSA as the local search algorithm). Next the behavior of our algorithm is tested on standard data sets. We show how MetaMax can be applied for tuning the parameters of machine learning algorithms: a classification task is solved by a neural network, and the parameters of the training algorithm (back-propagation) are fine-tuned by MetaMax combined with SPSA. MetaMax is also applied to boost the performance of $k$-means clustering. At the end of the section, we compare the results of the experiments to the theoretical bounds obtained in Section 4.2.

In all the experiments, in accordance with our simplifying assumptions introduced in Section 3, the main difference between the individual runs of the particular local search algorithm is their starting point. Obviously, more general diversification techniques exist: for example, the parameters of the local search algorithm could also vary from instance to instance (including running instances of different local search algorithms, where a parameter would select the actually employed search algorithm), and the initialization (starting point and parametrization) of a new instance could also depend on the results delivered by existing instances. Although the METAMAX strategies could also be applied to these more general scenarios, their behavior can be better studied in the simpler scenario; hence, our experiments correspond only to this setup.

---

6. For example, the relative difference between the average error $e_{\text{MetaMax}(\infty)}$ of MetaMax($\infty$) and $e_{\text{MetaMax}}$ of MetaMax in optimizing the parameters of a multi-layer perceptron for learning the *letter* data set (see Section 5.1 and especially Figure 6, right for more details) was 0.033 with a standard deviation of 0.06 (averaged over 1000 experiments), where the relative difference is defined as $\left| e_{\text{MetaMax}(\infty)} - e_{\text{MetaMax}} \right| / \max(e_{\text{MetaMax}(\infty)}, e_{\text{MetaMax}})$.

## 5.1 Optimizing Parameters with SPSA

In this section we compare the two versions of the METAMAX algorithm with six multi-start strategies, including three with a constant and three with a variable number of algorithm instances. The strategies are run for a fixed $T$ time steps, that is, the target function can be evaluated $T$ times, together by all local search instances (note that several reference strategies use $T$ as a parameter).

We used SPSA (Simultaneous Perturbation Stochastic Approximation; Spall, 1992) as the base local search algorithm in all cases. SPSA is a local search algorithm with a sampling function that uses gradient descent with a stochastic approximation of the derivative: at the actual location $X_t = (X_{t,1}, \ldots, X_{t,d})$, SPSA estimates the $l$th partial derivative of $f$ by

$$\tilde{f}_{t,l}(X_{t,l}) = \frac{f(X_t + \phi_t B_t) - f(X_t - \phi_t B_t)}{2\phi_t B_{t,l}},$$

where the $B_{t,l}$ are i.i.d. Bernoulli random variables that are the components of the vector $B_t$, and then uses the sampling function $s_t(X_t) = X_t + a_t \tilde{f}_t(X_t)$ to choose the next point to be sampled, that is,

$$X_{t+1,l} = X_{t,l} + a_t \tilde{f}_{t,l}(X_{t,l})$$

for $l = 1, \ldots, d$ ($\phi_t$ and $a_t$ are scalar parameters).

In the implementation of the algorithm we have followed the guidelines provided in (Spall, 1998), with the gain sequence $a_t = a/(A + t + 1)^\alpha$, and perturbation size $\phi_t = \phi/(t + 1)^\gamma$, where $A = 60$, $\alpha = 0.602$ and $\gamma = 0.101$. The values of $a$ and $\phi$ vary in the different experiments; they are chosen heuristically based on our experience with similar problems (this should not cause any problem here, as the goal of the experiments is not to provide fast solutions of the global optimization problems at hand but to demonstrate the behavior of the multi-start algorithms to be compared). In addition to the two evaluations required at the perturbed points, we also evaluate the function at the current point $X_t$. The starting point is chosen randomly, and the function is evaluated first at this point.

The six reference algorithms the METAMAX(K) and METAMAX algorithms are compared to the following:

UNIF: This algorithm selects from a constant number of instances of SPSA uniformly. In our implementation the instance $I_t = t \mod K$ is selected at time $t$, where $K$ denotes the number of instances.

THRASC: The Threshold Ascent algorithm of Streeter and Smith (2006b). The algorithm begins with selecting each of a fixed number of instances once. After this phase at each time step $t$ THRASC selects the best $s$ estimates produced so far by all algorithm instances $A_i, i = 1, \ldots, K$ in all the previous time steps, and for each $A_i$ it counts how many of these estimates were produced by $A_i$. Denoting the latter value by $S_{i,t}$, at time $t$ the algorithm selects the instance with index $I_t = \arg\max_i U(S_{i,t}/n_{i,t}, n_{i,t})$, where $n_{i,t}$ is the number of times the $i$th instance has been selected up to time $t$,

$$U(\mu, n) = \mu + \frac{\alpha + \sqrt{2n\mu\alpha + \alpha^2}}{n}$$

and $\alpha = \ln(2TK/\delta)$. $s$ and $\delta$ are the parameters of the algorithm, and in the experiments the best value for $s$ appeared to be 100, while $\delta$ was set to 0.01. We note that Threshold Ascent has been developed for the maximum $K$-armed bandit problem; nevertheless, it provides sufficiently good performance in our setup to test it in our experiments.

RAND: The random search algorithm. It can be seen as running a sequence of SPSA algorithms such that each instance is used for exactly one step, which is the evaluation of the random starting point of the SPSA algorithm.

LUBY: The algorithm based on the work of Luby et al. (1993). This method runs several instances of SPSA sequentially after each other, where the $i$th instance is run for $t_i$ steps, with $t_i$ defined by

$$t_i = \begin{cases} 2^{k-1}, & \text{if } i = 2^k - 1 \\ t_{i-2^{k-1}+1}, & \text{if } 2^{k-1} \le i < 2^k - 1 \end{cases}$$

The above definition produces a schedule such that from the first $2^k - 1$ algorithm instances one is run for $2^{k-1}$ steps, two for $2^{k-2}$ steps, four for $2^{k-3}$ steps, and so on.

EE-UNIF: This algorithm is an instance of explore-and-exploit algorithms. For the first $T/2$ steps the UNIF algorithm is used for exploration, and, subsequently, in the exploration phase, the SPSA instance that has achieved the highest value in the exploration phase is selected.

EE-LUBY: This algorithm is similar to EE-UNIF, except that LUBY is used for exploration.

Both versions of the METAMAX algorithm were tested. Motivated by the fact that SPSA is known to converge to a global optimum exponentially fast if $f$ satisfies some restrictive conditions (Gerencsér & Vágó, 2001), we chose a $h_r(n)$ that decays exponentially fast. To control the exploration of the so far suboptimal algorithm instances, we allowed $h_r(n)$ to be a time-varying function, that is, it changes with $t_r$, the total number of function calls to evaluate $f$ (or equally, the total number of steps taken) by all algorithms so far. Thus, at round $r + 1$ we used

$$h_r(n) = e^{-n/\sqrt{t_r}} \tag{27}$$

(note that we used the time-varying version of $h_r$ also in the case of METAMAX(K) – the latter can easily be extended to this situation, but this has been omitted to simplify the presentation).

For the algorithms with a fixed number of local search instances (METAMAX(K), UNIF, EE-UNIF, and THRASC), the number of instances $K$ was set to 100 in the simulations, as this choice provided reasonably good performance in all problems analyzed.

The multi-start algorithms were tested using two versions of a synthetic function, and by tuning the parameters of a learning algorithm on two standard data sets.

The synthetic function was a slightly modified[7] version of the Griewank function (Griewank, 1981):

$$f(x) = \prod_{l=1}^{d} \cos \frac{2\pi x_l}{\sqrt{l}} - \sum_{l=1}^{d} \frac{4\pi^2 x_l^2}{100}$$

where $x = (x_1, \ldots, x_d)$ and the $x_l$ were constrained to the interval $[-1, 1]$. We show the results for the 2-dimensional and the 10-dimensional cases.

The parameters of SPSA were $a = 0.05$ and $\phi = 0.1$ for the 2-dimensional case, and $a = 0.5$ and $\phi = 0.1$ for the 10-dimensional case. The performance of the search algorithms were measured by the error defined as the difference between the maximum value of the function (in this case 1) and the best result obtained by the search algorithm in a given number of steps. The results for the above multi-start strategies for the two- and the 10-dimensional test functions are shown in Figure 5. Each error curve is averaged over 10,000 runs, and each strategy was run for 100,000 steps (or iterations). One may observe that in both cases the two versions of the METAMAX algorithm converge the fastest. THRASC is better than UNIF, while LUBY seems fairly competitive with these two. The two explore-and-exploit-type algorithms (EE-UNIF and EE-LUBY) have similar performance on the 2-dimensional function, and clearly better than their 'non-exploiting' base algorithms, but on the 10-dimensional function their behavior is somewhat pathological in the sense that for low values of $T$ their performances are the best among all algorithms, but with increasing $T$, the error actually increases such that their respective base algorithms achieve smaller errors for some values of $T$. The random search seems an option only for the 2-dimensional function. Similar results were obtained for dimensions between 2 and 10. The pathological behavior of the explore-and-exploit algorithms start to appear gradually starting from the 5-dimensional function, and it is pronounced from 8 dimensions onwards. Limited experimental data obtained for higher dimensions up to 100 (averaged over only a few hundred runs) shows that the superiority of METAMAX is preserved for high-dimensional problems as well.

The reason for the pathological behavior of the explore-and-exploit strategies (i.e., why the error curves are not monotone decreasing in the number of iterations) can be illustrated as follows. Assume we have two SPSA instances, one converging to the global optimum and another one converging to a suboptimal local optimum. Assume that in the first few steps the optimal algorithm gives better result, then the suboptimal algorithm takes over and reaches its local maximum, while if the algorithms are run even further, the optimal algorithm beats the suboptimal one. If the exploration is stopped in the first or the last regime, an explore-and-exploit algorithm will choose the first, optimal local search instance, whose performance may get to quite close to the global optimum in the exploitation phase (even if it is stopped in the first regime). If the exploration is stopped in the middle regime, the suboptimal search instance will be selected for exploitation, whose performance may not even get close to the global optimum. In this scenario, the error after the exploitation phase (i.e. at the end) is lower if $T$ is small, and increases with higher values of $T$. Decrease in the error with increasing $T$ is only assured when the optimal instance converges in the exploration phase past all suboptimal local optima, which results in selecting the optimal local search instance for exploitation. In the above scenario the error will decrease fast initially, then increase for a while and then it may decrease again till it converges to 0, which is quite similar to what we observe in Figure 5, right. This pathological behavior becomes more transparent if there are many local search algorithms, as the length of the exploitation phase scales with the number of local search instances if the length of the exploration for each instance is kept fixed. Analyzing the experimental data shows that more complex versions of the scenario outlined above have occurred in the simulations and are the main cause of the observed pathological behavior (the non-monotonicity of the error curves).

---

7. The modification was made in order to have more significant differences between the values of the function at the global maximum and at other local maxima.

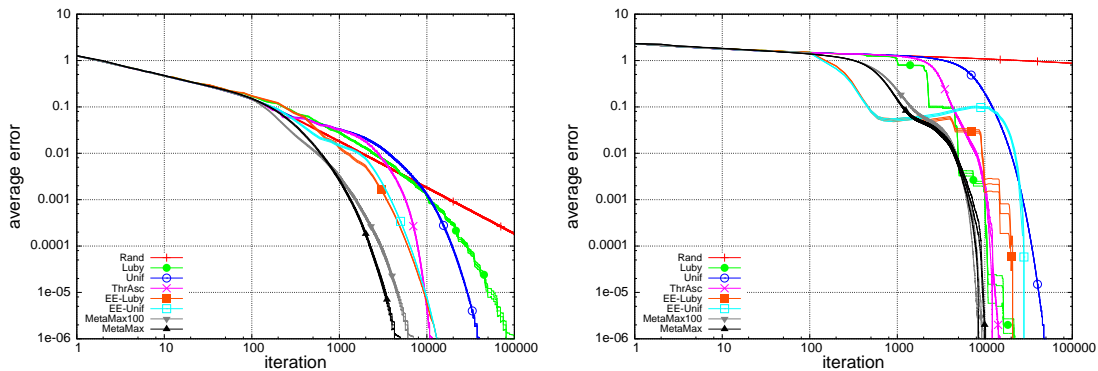

Figure 5: The average error for the multi-start strategies on the 2-dimensional (left) and 10-dimensional (right) modified Griewank function. 99% confidence intervals are shown with the same color as the corresponding curves. Note that most of the intervals are very small.

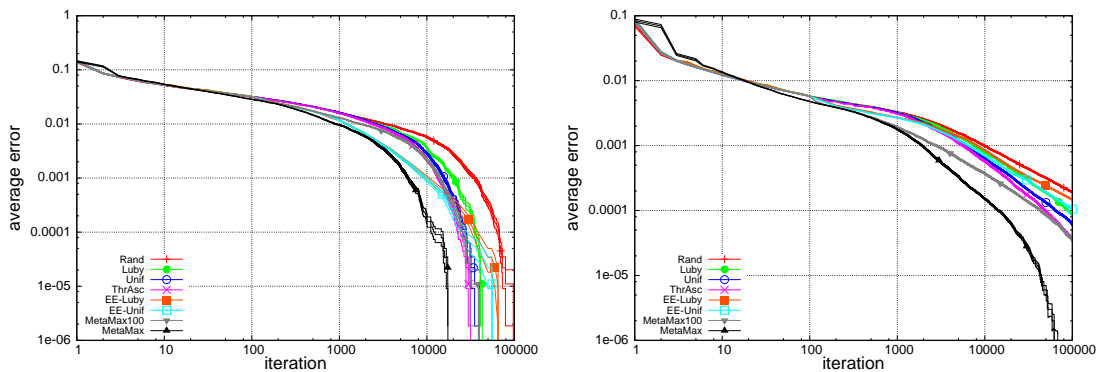

Figure 6: The average error for the multi-start strategies on tuning the parameters of Multilayer Perceptron for the *vehicle* data set (left) and the *letter* data set (right). 99% confidence intervals are also shown with the same color as the corresponding curves.

For tuning the parameters of a learning algorithm, we have used two standard data sets from the UCI Machine Learning Repository (Asuncion & Newman, 2007): *vehicle* and *letter*, and the Multilayer Perceptron learning algorithm of Weka (Witten & Frank, 2005) (here the back-propagation algorithm is used in the training phase). Two parameters were tuned: the learning rate and the momentum, both in the range of $[0, 1]$. The size of the hidden layer for the Multilayer Perceptron was set to 8, while the number of epochs to 100. The parameters of the SPSA algorithm were $a = 0.5$ and $\phi = 0.1$, the same as for the 10-dimensional Griewank function (as in the previous experiment, the parameters were chosen based on our experience). The rate of correctly classified items on the test set for *vehicle* using Multilayer Perceptron with varying values of the two parameters is shown in Figure 7, with the highest rate being 0.910112. Similarly, the classification rate for *letter* is shown in Figure 8, with the highest rate being 0.7505.

The error rates of the optimized Multilayer Perceptron on the data sets *vehicle* and *letter* are shown in Figure 6, when the parameters of the learning algorithm were tuned by the multi-start strategies above. The error in these cases is the difference between the best classification rate that can be obtained (0.910112 and 0.7505, respectively) and the best classification rate obtained by the multi-start strategies in a given number of steps. The results shown are averaged over 1,000 runs. We observe that the METAMAX algorithm (with an increasing number of algorithm instances) converged fastest in average, the three strategies with a fixed number of algorithm instances had nearly identical results, LUBY (and its explore-and-exploit variant) was slightly worse than these, and the random search was the slowest, although it performed not nearly as badly as for the synthetic functions. The reason why the random search had relatively better performance (relative to those that used SPSA) could be twofold: (i) large parts of the error surface offer fairly small error, and (ii) the error surface is less smooth, and therefore SPSA is less successful in using gradient information. The explore-and-exploit variants performed well on the *vehicle* data set initially, but their performance worsened for larger values of $T$ (compared to METAMAX, and to other algorithms to some extent). This, coupled with the observation for Figure 5, right would suggest that explore-and-exploit variants are more competitive for small values of $T$, despite their asymptotic guarantees.

In summary, the METAMAX algorithm (with an increasing number of algorithm instances) provided by far the best performance in all tests, usually requiring significantly fewer steps to find the optimum than the other algorithms. E.g., for the *letter* data set only the METAMAX algorithm found the global optimum in all runs in 100,000 time steps. We can conclude that METAMAX converged faster than the other multi-start strategies investigated in all four test cases, with a notable advantage on the difficult surfaces (at least from a gradient-based optimization viewpoint) induced by the classification tasks.

## 5.2 $k$-Means Clustering

In this section we consider the problem of partitioning a set of $M$ $d$-dimensional real vectors $x_j \in \mathbb{R}^d, j = 1, \ldots, M$ into $N$ clusters, where each cluster $S_i$ is represented by a center (or reconstruction) point $c_i \in \mathbb{R}^d$, $i = 1, \ldots, N$. The cost function to be minimized is the sum of distances $\rho(x, c_i)$ from the data points to the corresponding centers, that is, we want to minimize $\sum_{i=1}^{N} \sum_{x \in S_i} \rho(x, c_i)$. There are two necessary conditions for optimality (see, e.g., Linde, Buzo, & Gray, 1980; Gersho & Gray, 1992): for all $i = 1, \ldots, N$,

$$S_i = \{x : \rho(x, c_i) \leq \rho(x, c_j) \text{ for all } j = 1, \ldots, N\} \tag{28}$$

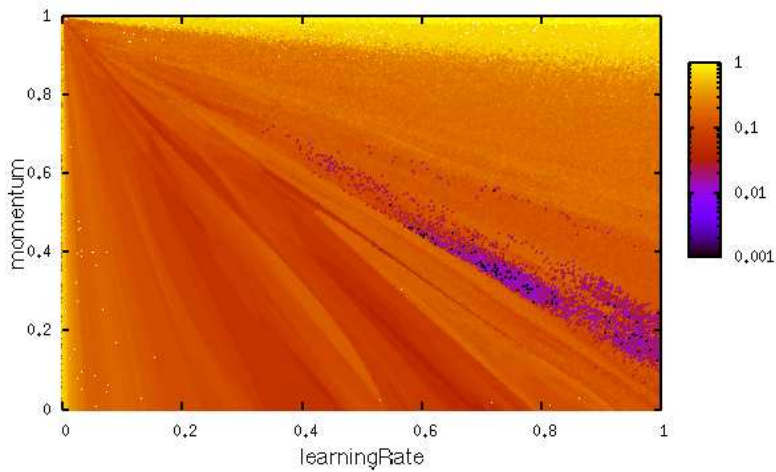

Figure 7: Classification rate on the *vehicle* data set. The rates are plotted by subtracting them from 0.911112 and thus global optima are at the scattered black spots corresponding to a value equal to 0.001.

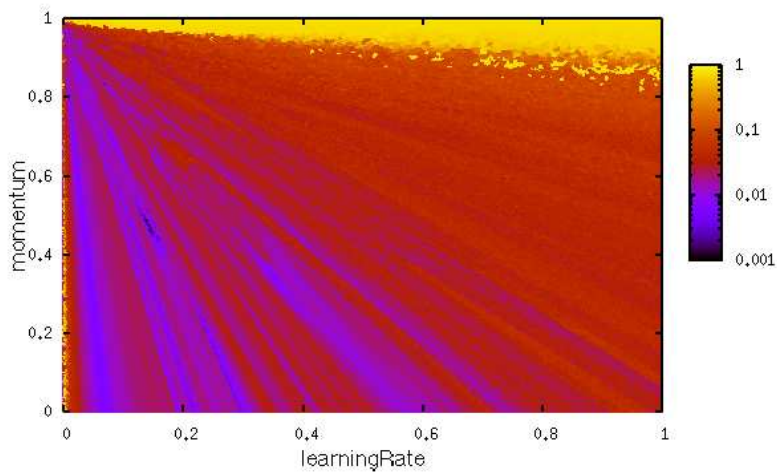

Figure 8: Classification rate on the *letter* data set. The rates are plotted by subtracting them from 0.7515 and thus global optima are at the scattered black spots corresponding to a value equal to 0.001.

(with ties broken arbitrarily) and

$$c_i = \operatorname*{argmin}_{c \in \mathbb{R}^d} \sum_{x \in S_i} \rho(x, c). \tag{29}$$

A usual choice for $\rho$ is the squared Euclidean distance, in which case $c_i = \frac{\sum_{x \in S_i} x}{|S_i|}$.[8] According to the above necessary conditions, the $k$-means algorithm (or Generalized-Lloyd algorithm, see, e.g., Linde et al., 1980; Gersho & Gray, 1992) alternates between partitioning the data set according to (28) with the centers being fixed, and recomputing the centers by (29) while the partitioning is kept fixed. It can easily be seen that the cost (or error) cannot increase in any of the above steps, hence the algorithm converges to a local minimum of the cost function. In practice, the algorithm stops when there is no (or insufficient) decrease in the cost function. However, the $k$-means algorithm is often trapped in a local optimum, whose value is influenced by the initial set of centers. As with SPSA, restarting $k$-means with a different initialization may result in finding the global optimum. We consider two initialization techniques: the first, termed K-MEANS, chooses the centers uniformly at random from the data points; the second, K-MEANS++ (Arthur & Vassilvitskii, 2007) chooses an initial center uniformly at random from the data set, and then chooses the further centers from the data points with a probability that is proportional to the distance between the data point and the closest center already selected.

The $k$-means algorithm usually terminates after a relatively small number of steps, and thus multi-start strategies with bounded number of instances would run out of active local search algorithms, and therefore do not appear particularly attractive. However, this is a natural domain to consider the strategy that starts a new instance, when the previous has finished. This strategy will be referred subsequently as SERIAL. For the above mentioned considerations, we test only METAMAX, the variant of our algorithms applicable for an unbounded number of instances.[9] As in the experiments with SPSA, we used $h_r$ as in (27). Note that some theoretical results indicate that $k$-means may converge at an exponential rate (in particular, Kieffer, 1982 showed that the rate of convergence is exponential for random variables with log-concave densities in 1-dimension provided that the logarithm of the density is not piecewise affine).

Two multi-start strategies, SERIAL and METAMAX were tested for the data set *cloud* from the UCI Machine Learning Repository (Asuncion & Newman, 2007). The data set was employed by Arthur and Vassilvitskii (2007) as well. The number of clusters was set to ten. The performance of the multi-start strategies is defined as the difference between the smallest cost function obtained by the strategy in a given number of steps and the smallest cost seen in any of the experiments (5626.6357). The results averaged over 1,000 runs are plotted in Figure 9. With both initialization methods the METAMAX strategy converges faster then the SERIAL strategy. We note that for this data set, K-MEANS++ with its more clever initialization procedure yields faster convergence than the standard K-MEANS with its uniform initialization, which is consistent with the results presented by Arthur and Vassilvitskii (2007).

---

8. An extension to clustering random variables is well-known and straightforward, but is omitted here because in this paper we only consider clustering finite data sets.

9. Note that in the METAMAX algorithm we have to do the practical modification that if a local search algorithm has terminated then it will not be chosen anymore. This clearly improves the performance as an algorithm is not chosen anymore when no improvement can be observed.

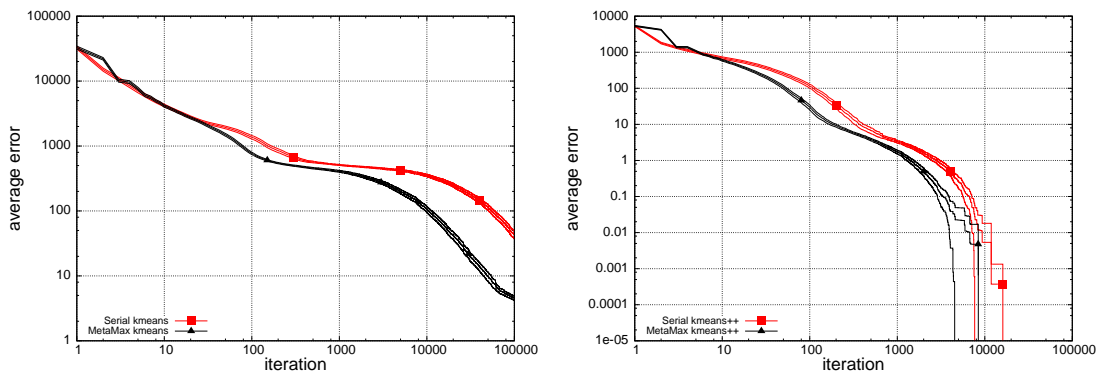

Figure 9: The average error for the multi-start strategies with k-means (left) and k-means++ (right). 99% confidence intervals are shown with the same color as the corresponding curves.

## 5.3 Practical Considerations

In all experiments with the MetaMax algorithm presented above, we observed that the number of algorithm instances $r$ (shown in Figure 10) grows at a rate of $\Omega(t_r/\ln t_r)$ (recall that $t_r$ is the total number of function calls to evaluate $f$, or the total number of steps, by all algorithm instances by the end of round $r$). On the other hand, in the derivation of the theoretical bounds (see Theorem 15 and Theorem 16) we used a bound $r \geq \Omega(\sqrt{t_r})$. In contrast to the quadratic penalty suggested by Theorem 16, plugging the $\Omega(t_r/\ln t_r)$ estimate of $r$ into the theorem we would find that only a logarithmic factor more calls to evaluate $f$ (total number of steps) are needed to achieve the performance of a search algorithm started from the attraction region of the optimum.

Finally, perhaps the main practical question concerning the MetaMax family of multi-start algorithms is to decide when to use them. As a rule of thumb, we have to say that there should be a sufficiently large performance difference between an "average" run of the local search algorithm and the best one. Clearly, if a single local search produces an acceptable result then it is not worth the effort to run several instances of the local search, especially not with a complicated schedule. In many real problems it is often the case that it is relatively easy to get close to the optimum, which may be acceptable for some applications, but approaching the optimum with greater precision is hard; if the latter is of importance, the MetaMax algorithm and its variants may be very useful. Last, one may wonder about the computational costs of our algorithms. As it was discussed before, we consider the case when the evaluation of the target function is very expensive: this is clearly not the case for the Griewank function, which is only used to demonstrate basic properties of the algorithm, but holds for many of the optimization problems in practice, including all other experiments considered in this paper. In these problems the function evaluation is indeed expensive (and depends on the available data), while the overhead introduced by the MetaMax algorithms depends on the number of rounds. For the MetaMax(K) algorithm we have to find the upper convex hull of a set of $K$ points in each round; in the worst case this can take as long as $O(K^2)$ calculations, but in practice this is usually much cheaper, as the upper convex hull is determined by the point that corresponds to the actually best estimate and by the point that corresponds to the least used algorithm, which requires only $O(K)$ computations, or even less, if some special ordering tricks are introduced. Since the target function $f$ is evaluated at least twice in each round, on average at most $O(K^2)$ computational overhead is needed for each evaluation of $f$ in the worst case, which is practically reduced to $O(K)$, or even less. Similar considerations hold for the METAMAX($\infty$) and the METAMAX algorithms, resulting in an average $O(r^2)$ worst-case overhead for each call to $f$ (in $r$ rounds), which is closer to $O(r)$ or even less in practice. In all examples we considered (apart of the case for the Griewank function), this amount of overhead has been negligible relative to the computational resources needed to evaluate $f$ at a single point.

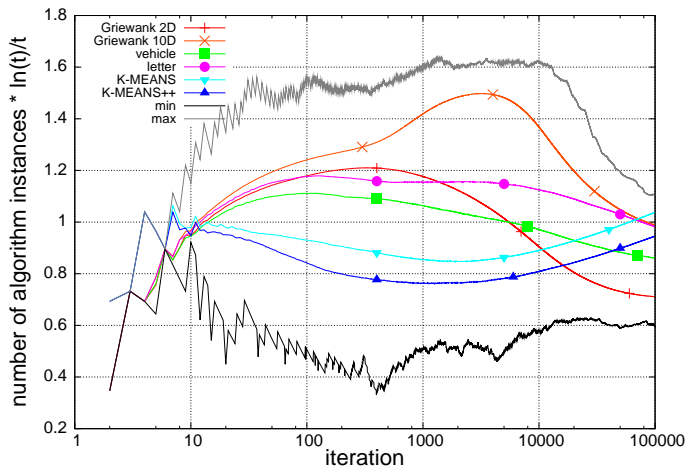

Figure 10: Number of algorithm instances ($r$) for METAMAX. The average number of instances are shown on the six benchmarks: Griewank function (2- and 10-dimensional), parameter tuning of Multilayer Perceptron (on the *vehicle* and on the *letter* data set), and clustering with K-MEANS and K-MEANS++. The maximum and minimum number of instances over all runs of all benchmarks are also shown. One can notice that for larger values of $t_r$, $0.45 t_r / \ln t_r \le r \le 1.65 t_r / \ln t_r$.

## 6. Conclusions

In this paper we provided multi-start strategies for local search algorithms. The strategies continuously estimate the potential performance of each algorithm instance in an optimistic way, by supposing a convergence rate of the local search algorithms up to an unknown constant, and in every phase resources are allocated to those instances that could converge to the optimum for a particular range of the constant. Three versions of the algorithm were presented, one that is able to follow the performance of the best of a fixed number of local search algorithm instances, and two that, with gradually increasing the number of the local search algorithms, achieve global consistency. A theoretical analysis of the asymptotic behavior of the algorithms was also given. Specifically, under some mild conditions on the function to be maximized (e.g., the set of the values of the local maxima is not dense at the global maximum), our best algorithm, MetaMax, preserves the performance of a local search algorithm for the original function class with at most a quadratic increase in the number of times the target function needs to be evaluated (asymptotically). Simulations demonstrate that the algorithms work quite well in practice.

While the theoretical bound suggests that the target function has to be evaluated a quadratic factor more times to achieve the performance of a search algorithm that is started from the attraction region of the optimum, in the experiments we found only a logarithmic penalty. It is not clear whether this difference is the result of our slightly conservative (asymptotic) analysis or the choice of the experimental settings. Also, a finite sample analysis of the algorithm is of interest, as the experiments indicate that the MetaMax algorithm provides good performance even for a relatively small number of steps taken by the local search algorithms, in the sense that it provides a speed-up compared to other approaches even if the number of times the target function can be evaluated (i.e., the total number of steps that can be taken by all algorithms together) is relatively small. Finally, future work is needed to clarify the connection between the convergence rate of the optimal algorithms ($g_\delta$) and the function $h_r$ used in the exploration.

## Acknowledgments

The authors would like to thank the anonymous referees for their numerous insightful and constructive comments. This research was supported in part by the Mobile Innovation Center of Hungary, by the National Development Agency of Hungary from the Research and Technological Innovation Fund (KTIA-OTKA CNK 77782), and by the PASCAL2 Network of Excellence (EC grant no. 216886). Parts of this paper were presented at ECML 2009 (Kocsis & György, 2009).

## Appendix A. Proof of Lemma 1

Fix $\delta \in (0, 1)$ and let $U_n = \hat{f}^* - f(\widehat{X}_n)$. Since $U_n \to 0$ almost everywhere on $E$, Egoroff's theorem (see, e.g. Ash & Doléans-Dade, 2000) implies that there is an event $E_\delta \subseteq E$ with $1 - P(E_\delta) < \delta$ such that $U_n \to 0$ uniformly almost everywhere on $E_\delta$. The second part of the lemma follows from the definition of uniform convergence. $\qquad\square$